\documentclass[journal]{IEEEtran}

\usepackage{cite}
\usepackage{graphicx} 
\usepackage{algorithm}
\usepackage{algorithmic}
\usepackage[utf8]{inputenc}
\usepackage[T1]{fontenc}
\usepackage[english]{babel}
\usepackage{enumerate}
\usepackage{amsmath}
\usepackage{amssymb}
\usepackage{longtable}
\usepackage{pdfpages}
\usepackage{color}

\def\I{\ensuremath{\mathbf{I}}}

\def\y{\ensuremath{\mathbf{y}}}
\def\v{\ensuremath{\mathbf{v}}}
\def\x{\ensuremath{\mathbf{x}}}
\def\b{\ensuremath{\mathbf{b}}}

\def\d{\ensuremath{\mathbf{d}}}

\def\H{\ensuremath{\mathbf{H}}}

\def\R{\mathbb{R}}

\def\z{\mathbf{z}}
\def\a{\mathbf{a}}
\def\b{\mathbf{b}}
\def\u{\mathbf{u}}
\def\0{\mathbf{0}}

\DeclareMathOperator*{\argmin}{arg\,min}
\def\prox{{\textbf{prox}}}

\newcommand{\citet}[1]{{\cite{#1}}}

\newcommand{\cor}[1]{{\textcolor{black}{#1}}}

\newcommand{\rrev}[1]{{\textcolor{black}{#1}}}
\newcommand{\remii}[1]{{\textcolor{black}{#1}}}
\newcommand{\modif}[1]{{\color{black} #1}}

\newcommand{\GGCorr}[2]{{\color{black}#2}}

\newtheorem{proposition}{Proposition}

\newtheorem{lemma}{Lemma}
\newtheorem{definition}{Definition}

\begin{document}

\title{DC Proximal Newton for  Non-Convex Optimization Problems} 
\author{ A. Rakotomamonjy, R. Flamary, G. Gasso 
 \thanks{This work has been partly supported by the French ANR (09-EMER-001, 12-BS02-004 and 12-BS03-003).}
\thanks{
 \newline AR is with LITIS EA 4108, Universit\'e de Rouen, France. alain.rakoto@insa-rouen.fr
 \newline RF is with Lagrange laroratory, Universit\'e de Nice
 Sophia-Antipolis, CNRS, Observatoire de la C\^ote d'Azur,  F-06304
 Nice, France. remi.flamary@unice.fr
 \newline GG is with LITIS EA 4108, INSA de Rouen, France. gilles.gasso@insa-rouen.fr }
}

\maketitle

 \begin{abstract}
 We introduce a novel algorithm for solving learning
  problems where both the loss function and the regularizer are
  non-convex but belong to the class of
   difference of convex (DC) functions.
Our contribution is a new general purpose  proximal
 Newton algorithm that is able to deal with such a situation. The
 algorithm consists in obtaining a descent direction from an
 approximation of the loss function and then in performing a line search to
 ensure sufficient descent. A theoretical analysis is provided showing that the
  iterates of the proposed algorithm {admit} as limit points stationary points of the
   DC objective function. Numerical experiments show that our approach
 is more efficient than current state of the art for a problem with a
 convex loss function and a non-convex regularizer. We have also
 illustrated the  benefit of our algorithm in high-dimensional
 transductive learning problem where both loss function and
 regularizers are non-convex.
 \end{abstract}


\IEEEpeerreviewmaketitle

\section{Introduction}
In many real-world application domains such as computational
biology, finance or text mining,  datasets considered for
learning  prediction models are routinely large-scale and
high-dimensional \GGCorr{}{raising the issue of model complexity control}. One way for dealing with such kinds  of dataset is
to learn sparse models. Hence, a very large amount of recent
works in machine learning, statistics and signal processing have
addressed optimization problems related to sparsity issues.

One of the most popular algorithm for achieving sparse models is the
Lasso algorithm \cite{Tibshrani_Lasso_1996} also known as the Basis
pursuit algorithm \cite{chen_bpjournal} in the signal processing
community. This algorithm actually applies $\ell_1$-norm
regularization to the learning model.  The choice of the $\ell_1$ norm
comes from its appealing properties which are convexity, continuity
and its ability to produce sparse or even the sparsest model in some cases
{owing to its non-differentiability at zero}  
 \cite{li10:_two_condit_equiv_norm_solut,donoho06:_for}.
Since these seminal works,
several efforts have been devoted to the development of efficient
algorithms for solving learning problems that consider
sparsity-inducing regularizers
\cite{shevade03:_simpl_and_effic_algor_for,beck09:_fast_iterat_shrin_thres_algor,yuan2012improved,bach11:_convex}.  However, $\ell_1$ regularizer presents
some drawbacks such as its inability, in certain situations to
retrieve the true relevant variables of a model
\cite{zou_adaptive_lasso_2006,Fan_LI_scad_2001}. Since the
$\ell_1$-norm regularizer is a continuous and convex surrogate of the
$\ell_0$ pseudo-norm, other kinds of regularizer which abandon the
convexity property, have been analyzed by several authors and they
have been proved to achieve better statistical property. Common
non-convex and {non-differentiable} regularizers are the SCAD regularizer \cite{Fan_LI_scad_2001}, the $\ell_p$
regularizer \cite{KnightFuAsymptotics}, the capped-$\ell_1$  and the \textit{log} penalty \cite{CandesReweighted2008}. These regularizers have been 
frequently used for feature selections or for 
obtaining sparse models \cite{laporte13:_noncon_regul_featur_selec_rankin,gasso09:_recov_spars_signal_with_certain,CandesReweighted2008}. 

While being statistically appealing, the use of these non-convex {and  non-smooth} regularizers
poses some challenging optimization problems. In this work, we propose
a novel efficient non-convex proximal Newton algorithm. Indeed, one of the most frequently
used  algorithms for solving $\ell_1$-norm regularized problem is the  proximal gradient algorithm \cite{combettes11:_proxim_split_method_in_signal_proces}. Recently, proximal Newton-type
methods have been introduced for solving composite optimization problems
involving the sum of a smooth and convex twice differentiable function and a 
non-smooth convex function (typically the regularizer)\cite{lee2012:_proximal,becker12:_newton}. These proximal Newton
algorithms have been shown to be substantially faster than their proximal gradient counterpart.  {Our objective is thus to go beyond
the state-of-the-art by proposing an efficient proximal Newton algorithm
that is able to handle machine learning problems where the loss function
is smooth and possibly non-convex and the regularizer is non-smooth and
non-convex. 
}

Based on this, we propose an {effficient} general proximal Newton method for
optimizing a composite objective function $f(\x)+h(\x)$ where both
functions $f$ and $h$ can be  non-convex and belong to a
large class of functions that can be decomposed as the difference of
two convex functions (DC functions) \cite{thi05:_dc_differ_convex_funct_progr,dinh97:_convex_analy_approac_dc_progr,akoa08:_combin_dc_algor_dcas_decom}. 
{In addition, we also allow $h(\x)$ to be non-smooth, which is
  necessary for sparsity promoting regularization}.
The proposed algorithm has a wide
range of applicability that goes far beyond the handling of non-convex
regularizers. Indeed, our global framework can genuinely deal with
non-convex loss functions that usually appear in learning problems.
To make concrete the DC Newton proximal approach, we illustrate the
relevance and the effectiveness of the novel algorithm by considering
a problem of sparse transductive logistic regression in which the
regularizer as well as the loss related to the unlabeled examples are
non-convex. As far as our knowledge goes, this is the first work that
introduces such a model and proposes an algorithm for solving
the related optimization problem. 
{In addition to this specific problem, many non-convex optimization
problems involving non-convex loss functions and non-convex and non-differentiable regularizers arise in machine learning \emph{e.g} dictionary learning \cite{jenatton10:_proxim_method_for_spars_hierar_diction_learn,rakotomamonjy12:_direc} or matrix factorization \cite{srebro05:_maxim_margin_matrix_factor} problems.
In addition, several works have recently shown that non-convex loss functions
such as the Ramp loss which is a DC function, lead to classifiers more
robust to outliers \cite{ertekin11:_noncon,collobert:2006}. 
We thus believe that the proposed framework is of general interest in machine
learning optimization problems involving this kind of losses and regularizers.}

The  algorithm we propose consists in two steps: first it seeks  a search direction and then it looks for a step-size in that
direction that minimizes the objective value. The originality and main novelty we brought in this work is that the
search direction is obtained by solving a subproblem which involves
both an approximation of the smooth loss function and the DC regularizer.
{ Note that while our algorithm for non-convex objective function is 
rather similar to the convex proximal Newton method,  non-convexity and 
non-differentiability  raise some technical issues when analysing
the properties of the algorithm.
Nonetheless, we prove several properties related to the  search direction and provide convergence analysis of the algorithm to a stationary point
of the related optimization problem. These properties are obtained
as non-trivial extension of the convex proximal Newton case.}
 Experimental studies show
the benefit of the algorithm in terms of running time while preserving or improving generalization performance compared to existing non-convex approaches.

\modif{
The paper is organized as follows. Section \ref{section:fram} introduces the
general optimization problem we want to address as well as
the  proposed DC proximal Newton optimization scheme. Details
on the implementation and discussion concerning related 
works are also provided. In Section \ref{sec:analysis}, an analysis of the properties
of the algorithm is given.
Numerical experiments on simulated and real-world data comparing our
approach to the existing methods are depicted in Section \ref{sec:expe}, while
Section \ref{sec:conclu} concludes the paper.
}

\section{DC proximal Newton algorithm}
\label{section:fram}
We are interested in solving the following optimization problem
\begin{equation}
  \label{eq:optprob}
  \min_{\x \in \R^d}\, F(\x) :=  f(\x)+h(\x)
\end{equation}
with the following assumptions concerning the functions 
$f$ and $h$.  $f$ is supposed  to be twice differentiable, 
lower bounded on $\R^d$ and  we suppose that there exists two convex functions
$f_1$ and $f_2$ such that $f$ can be written as a difference
of convex  (DC) functions $f(\x)= f_1(\x) - f_2(\x)$.
We also assume that $f_1$ verifies the $L$-Lipschitz gradient property
  \begin{equation*}
    \|\nabla f_1(\x)-\nabla f_1(\y)\|\leq L \|\x-\y\|\quad \forall \x,\y
    \in \text{dom} f_1.
  \end{equation*}
 \modif{The DC assumption on $f$ is not very 
restrictive since any differentiable function $f(\cdot)$ with a bounded
 Hessian matrix can be expressed as a difference of convex function \cite{Yuille_CCCP_2001}.}

The function $h$ is  supposed to be a lower-bounded, proper, lower semi-continuous and its restriction to its domain is continuous. We suppose that $h$ that can also be expressed  as   \begin{equation}
    \label{eq:h_dc}
    h(\x)=h_1(\x)-h_2(\x)
  \end{equation}
where $h_1$ and $h_2$ are both convex functions. 
{As discussed in the introduction, we focus our interest in
situations where $h$ is {non-convex and non-differentiable.
As such $h_1$ and $h_2$ are also expected to be non-differentiable.}
\modif{A large class of non-convex sparsity-inducing regularizers can be
expressed as a DC function as discussed in
\cite{gasso09:_recov_spars_signal_with_certain}. This includes  the classical SCAD regularizer, the $\ell_p$
regularizer, the capped-$\ell_1$  and the \textit{log} penalty as above-mentioned.
}
}

\modif{Note that those assumptions on $f$ and $h$ cover a broad class of optimization problems. Proposed approach can be applied for sparse linear model
estimation as illustrated in in Section \ref{sec:expe}. But more
general learning problems such as those using  overlapping nonconvex $\ell_p-\ell_1$ (with $p<1$) group-lasso as used in \cite{courty2014domain} can also be considered. Our framework also encompasses those of structured sparse dictionary learning or matrix factorization \cite{jenatton10:_struc_spars_princ_compon_analy,rakotomamonjy12:_direc}, sparse and low-rank matrix estimation \cite{richard2012estimation,deng2013low}, or maximum likekihood estimation of
graphical models \cite{Zhong_nips}, when the $\ell_1$ sparsity-inducing regularizer is replaced for instance by a more aggressive regularizer like the \emph{log} penalty or the SCAD regularizer.

}

\subsection{Optimization scheme}

For solving Problem (\ref{eq:optprob}) which
is a difference of convex functions optimization problem, we propose
a novel iterative algorithm which first looks for 
a search direction $\Delta \x$  and then updates
the current solution. Formally, the algorithm is
based on the 
iteration
$$
\x_{k+1}= \x_{k} + t_k \Delta \x_k
$$
where $t_k$ and $\Delta \x_k$ are respectively a step size
and the search direction. Similarly to 
the works of Lee et al. \cite{lee2012:_proximal}, the search  direction
is computed by minimizing a local approximation 
of the composite function $F(\x)$.
However, we show that by using a simple approximation
on $f_1$, $f_2$ and $h_2$, we are able to  handle the
non-convexity of $F(\x)$, resulting  in an
algorithm which is wrapped around a specific proximal Newton 
iteration.

For dealing with the non-convex situation, we define the search direction 
as the solution of the following 
problem 
\begin{eqnarray}\label{eq:iter_approx}
  \Delta \x_{k}=\argmin_{\Delta \x} \tilde f (\x_k + \Delta \x ) +  \tilde h(\x_k + \Delta \x )
\end{eqnarray}
where $\tilde f$ and $\tilde h$ are the following approximations
of respectively  $f$ and $h$ at $\x_k$. We define $\tilde f (\x)$ as
\begin{eqnarray}
  \label{eq:approx_f}
  \tilde f (\x)&=& f_1(\x_k) + \nabla f_1(\x_k)^\top(\x -\x_k) \\ \nonumber
 &&+~   \frac{1}{2}(\x -\x_k)^\top \H_k(\x -\x_k)  \\\nonumber
&& -~f_2(\x_k) - \z_{f_2}^\top(\x-\x_k) \nonumber
\end{eqnarray}
\GGCorr{where $\H_k$ is a positive definite matrix and $\z_{f_2}= \cor{\nabla}f_2(\x_k)$.}{where $\z_{f_2}= \cor{\nabla}f_2(\x_k)$ and $\H_k$ is any positive definite approximation of the Hessian matrix of $f_1$ at current iterate.}
We also consider 
\begin{equation}
  \label{eq:approx_h}
  \tilde h (\x)=h_1(\x)-h_2(\x_k)- \z_\cor{h_2}^\top(\x -\x_k) 
\end{equation}
where $\z_\cor{h_2} \in \partial h_2(\x_k) $, with the latter being
the sub-differential of $h_2$ at $\x_k$. 

Note that the first three summands in Equation (\ref{eq:approx_f}) form a quadratic approximation
of $f_1(\x)$ whereas the terms in the third line of 
Equation (\ref{eq:approx_f}) is a linear approximation
of $f_2(\x)$. In the same spirit, $\tilde h$ is
actually a majorizing function of $h$ since we have linearized
the convex function $h_2$ and $h$ is a difference of convex
functions. 

We are now in position
to provide the proximal expression of the search direction. 
Indeed, Problem (\ref{eq:iter_approx}) can be rewritten as 
\begin{equation}\label{eq:search}
  \argmin_{\Delta \x} \frac{1}{2} \Delta \x^\top \H_k\Delta \x  +h_1(\x_k +
\Delta \x)  + {\v_k}^\top\Delta \x 
\end{equation}
with $\v_k= \nabla f_\cor{1}(\x_k)-\z_{f_2} - \z_{h_2}$.
After some algebras {given in the appendix} and involving optimality conditions
of a proximal Newton operator, we can show
that
\begin{equation}\label{eq:searchprox}
 \Delta \x_k =  \prox_{h_1}^{\H_k}(\x_k - \H_k^{-1}\v_k) - \x_k
\end{equation}
with by definition \cite{combettes11:_proxim_split_method_in_signal_proces,lee2012:_proximal}
$$
\prox_{h_1}^{\H}(\x)= \arg\min_\y \frac{1}{2} \|\x - \y\|_{\H}^2 + h_1(\y)
$$
{where $\|\x\|_{\H}^2=\x^\top\H\x$ is the quadratic norm with metric $\H$.}
Interestingly, we note that the  non-convexity of the initial problem is taken  into
account only through the proximal Newton operator and its impact
on the algorithm, compared to the convex case, is minor
since it only modifies the argument of the operator through $\v_k$. 

Once the search direction is computed, the step size $t_k$ is backtracked
starting from $t_k=1$. Algorithm \ref{alg:algo} summarizes the main steps of the optimization
scheme. Some implementation issues are discussed hereafter while the
next section focuses on the convergence analysis.

\begin{algorithm}[t]
\caption{DC proximal Newton algorithm}
\label{alg:algo}
\begin{algorithmic}[1]
 \STATE Initialize $\x_0 \in dom \cor{F}$ 
 \STATE $k=0$  
 \REPEAT 
 \STATE compute $\z_{h_2} \in \partial h_2(\x_k)$ and $\z_{f_2} = \nabla f_2(\x_k)$
 \STATE update $\H_k$ (exactly or using a quasi-Newton approach)
\STATE $\v_k \leftarrow \nabla f_1(\x_k) - \z_{f_2} -\z_{h_2}$ 
 \STATE  $\Delta \x_k \leftarrow   \prox_{h_1}^{\H_k}(\x_k - \H_k^{-1}\v_k) - \x_k$
\STATE compute the stepsize $t_k$ through backtracking starting from $t_k=1$
\STATE $\x_{k+1}= \x_k + t_k \Delta{\x_k}$
\STATE $k \leftarrow k+1$
\UNTIL{convergence criterion is met}
\end{algorithmic}
\end{algorithm}

\subsection{Implementation's tricks of the trade}
The main difficulty and computational burden of our DC proximal Newton
algorithm resides in the computation of the search direction
$\Delta \x_k$. Indeed, the latter needs the computation of
the  proximal operator $\prox_{h_1}^{\H_k}(\x_k - \H_k^{-1}\v_k)$
which is equal to 
\begin{align} \label{eq:search2}
& \argmin_\y \underbrace{\frac{1}{2} \y^\top \H_k\y
\GGCorr{-}{+} \y^\top (\v_k - \H_k\x_k)}_{g(\y)} + h_1(\y) 
\end{align}

We can note that Equation \ref{eq:search2} represents a quadratic problem penalized by $h_1$. If
$h_1(\y)$ is a term which proximal operator can be cheaply computed 
then, one can consider proximal gradient algorithm or any other efficient algorithms for its resolution \cite{beck09:_fast_iterat_shrin_thres_algor,FigueiredoNowakGradProj2007}. 

In our case, we have considered a forward-backward (FB) algorithm \cite{combettes11:_proxim_split_method_in_signal_proces} initialized with the previous value of the optimal $\y$. 
Note that in order to have a convergence guarantee, the FB algorithm needs
a stepsize smaller than $\frac{2}{L}$ where $L$ is the Lipschitz gradient
of the quadratic function. Again computing $L$ can be expensive and in order
to increase the computational efficiency of the global algorithm, we have
chosen a strategy that roughly estimates $L$ according to the
equation
$$
\frac{\|\nabla g(\y) - \nabla g(\y^\prime\cor{)} \|_2}{\|\y - \y^\prime\|_2}
$$ 
In practice, we have found this heuristic to be slightly more efficient 
than an approach which computes the largest eigenvalue of $\H_k$ by
means of a power method \cite{golub1996matrix}. {Note that a
  L-BFGS approximation scheme has been used in the numerical
  experiments for updating the matrix $\H_k$ and its inverse.}

While the convergence analysis we provide in the next section
supposes that the proximal operator is computed exactly, in practice it
is more efficient to approximately solve the search direction problem,
at least for the early iterations. Following this idea, 
we have considered an adaptive stopping criterion for the proximal
operator subproblem.

\subsection{Related works} \label{subsec:related_works}

In the last few years, a large amount of works have been devoted to
the resolution of composite optimization  problem of the form given in Equation (\ref{eq:optprob}).{ We review the ones that are most similar
to ours and summarize the most important ones in Table \ref{tab:listmethods}.}

Proximal Newton algorithms have  recently been proposed by \cite{lee2012:_proximal} and \cite{becker12:_newton} for solving Equation (\ref{eq:optprob}) when 
both functions $f(\x)$ and $h(\x)$ are convex. While the algorithm
we propose is similar to the one of \cite{lee2012:_proximal},
our work  is strictly more general in the sense that we abandon the convexity
hypothesis on both functions. Indeed, our algorithm can handle both
convex and non-convex cases and boils down to the algorithm
of \cite{lee2012:_proximal} \cor{in the convex case}. 
{Hence, the main contribution that differentiates our work
to the work of Lee et al. \cite{lee2012:_proximal} relies on the extension
of the algorithm to the non-convex case and  the
theoretical analysis of the resulting algorithm. }

\begin{table}[t]

  \caption{Summary of related approaches according to how  $f(\x)$ and 
$h(\x)$ are decomposed in $f_1 - f_2$ and
$h_1 - h_2$.  \emph{cvx} and \emph{ncvx} respectively stands for
\emph{convex} and \emph{non-convex}. $-$ denotes that the method that does not handle
DC functions. The \emph{metric} column denotes the form of
the metric used in the quadratic approximation.  }
\label{tab:listmethods}
  \centering
  \begin{tabular}[h]{l|cccccc}
    &\multicolumn{2}{c}{f(\x)} &\multicolumn{2}{c}{h(\x)} & metric \\\hline  
Approach & $f_1$ & $f_2$ & $h_1$ & $h_2$ & $\H$     \\ \hline
proximal gradient \cite{combettes11:_proxim_split_method_in_signal_proces}& cvx & - & cvx & - & $\frac{L}{2}$\I    \\
proximal Newton \cite{lee2012:_proximal} & cvx & - & cvx & - & $\H_k$ \\
GIST \cite{gong2013jieping}& ncvx & - & cvx & cvx & $\frac{L}{2}$\I \\
SQP \cite{lu12:_sequen_convex_progr_method_class} & cvx & cvx & cvx & cvx & $\frac{L}{2}$\I \\
our approach  & cvx & cvx & cvx & cvx & $\H_k$ \\ \hline
  \end{tabular}
\end{table}

Following the interest on sparsity-inducing regularizers, there
has been a renewal of curiosity around non-convex optimization
problems \cite{CandesReweighted2008,laporte13:_noncon_regul_featur_selec_rankin}. Indeed, most statistically relevant sparsity-inducing
regularizers are non-convex \GGCorr{}{\cite{PoLing_NIPS2013}}. Hence, several \GGCorr{groups of research}{researchers}
have  proposed novel algorithms for handling these isssues.

We point out that linearizing the concave part in a DC program is
a crucial idea of DC programming and DCA that were introduced by Pham
Dinh Tao in the early eighties and have been extensively
developed since then \cite{thi05:_dc_differ_convex_funct_progr,dinh98:_dc_optim_algor_solvin_trust_region_subpr,dinh97:_convex_analy_approac_dc_progr}.
In this work, we have used this same idea in a proximal Newton framework.
However, our  algorithm is fairly different \GGCorr{to}{from} the
DCA \cite{dinh97:_convex_analy_approac_dc_progr} as we consider a single descent step at each iteration, as opposed to the DCA which needs a full optimization
of a  minimization problem at each iteration. {This DCA algorithm
has as special case, the convex concave procedure (CCCP) introduced
by Yuille et al. \cite{Yuille_CCCP_2001} and used for instance
by Collobert et. al \cite{tsvm_collobert_jmlr_2006} in a machine
learning context.}

This idea of linearizing the (possibly) non-convex part of Problem (\ref{eq:optprob}) for obtaining
a search direction can also be found in
Mine {and Fukushima} \cite{mine81}. However, in their case, the function to be linearized
is supposed to be smooth. The advantage of using a DC program, as in our case, is that the linearization trick can also be extended to non-smooth function.

The works that are mostly related to \GGCorr{our works}{ours} are those proposed
by \cite{gong2013jieping} and  \cite{lu12:_sequen_convex_progr_method_class}.
Interestingly, Gong et al. \cite{gong2013jieping} introduced a generalized iterative shrinkage
algorithm (GIST) that can handle optimization problems with DC regularizers 
for which proximal operators
can be easily computed. Instead, Lu \cite{lu12:_sequen_convex_progr_method_class}
solves  the same optimization problem in a different way.
As the non-convex regularizers \GGCorr{is}{are} supposed to be DC, he proposes to
solve a sequence of convex programs which at each iteration
minimizes $$
  \tilde f (\x) + h_1(\x)-h_2(\x_k)- \z_{h_2}^\top(\x -\x_k) 
$$
with 
$$
  \tilde f (\x) =  f_1(\x_k) + \nabla f_1(\x_k)^\top(\x -\x_k) +   \frac{{L}}{2}\|\x -\x_k\|^2 
$$
Note that  our framework subsumes the one of \GGCorr{Lu et al.}{Lu \cite{lu12:_sequen_convex_progr_method_class}} ({when considering unconstrained optimization problem}). Indeed,
we take into account a variable metric $\H_k$ into the proximal term. 
Thus, the approach of Lu can be deemed  a particular case of our method {where
$\H_k= {L\,} \I$ at all iterations of the algorithm. }
Hence, when $f(\x)$ is convex, we expect more efficiency 
 compared to the algorithms of \cite{gong2013jieping} and \cite{lu12:_sequen_convex_progr_method_class} owing to the variable metric $\H_k$ that has been introduced. 

Very recently, \GGCorr{\citet{chouzenoux13:_variab_metric_forwar_backw}}{Chouzenoux et al. \citet{chouzenoux13:_variab_metric_forwar_backw}} introduced
a proximal Newton-like algorithm for minimizing the sum of a twice differentiable
function and a convex function. They essentially consider
that the regularization term is  convex while the loss
function may be non-convex. Their work can thus be seen as an extension
of the one of \citet{sra11} to the variable metric $\H_k$ case. 
Compared to our work, \citet{chouzenoux13:_variab_metric_forwar_backw} 
do not impose a DC condition on the
function $f(\x)$. However, at each iteration, they need a quadratic surrogate
function at a point $\x_k$ that majorizes $f(\x)$. In our case, only the non-convex part is majorized through a simple linearization.

\section{Analysis}
\label{sec:analysis}
Our objective in this section is to show that
our algorithm is well-behaved and to prove
at which extents the iterates $\{\x_k\}$ converge to a
stationary point of Problem (\ref{eq:optprob}). 
 We first
characterize stationary points of Problem 
\ref{eq:optprob} with respects  to
$\Delta \x$ and then show 
that all limit points of the sequence
$\{\x_k\}$ generated by our algorithm are
 stationary points.\\

Throughout this work, we use the following definition of a stationary point. 

\begin{definition} A point $\x^\star$ is said to be a stationary
point of Problem (\ref{eq:optprob}) if 
$$
0 \in \nabla f_1(\x^\star) - \nabla f_2(\x^\star)  
+ \partial h_1(\x^\star) - \partial h_2(\x^\star)
$$
\end{definition}
Note that being a stationary point, as defined above, is a necessary
condition for a point $\x^\star$ to be a local minimizer of Problem
(\ref{eq:optprob}).

According to the above definition, we have the following lemma~:
\begin{lemma}
  Suppose $\H_\star \succ 0$, $\x^\star$ is a stationary point of  Problem (\ref{eq:optprob}) if and only if
$\Delta \x^\star=0$ with
\begin{equation}\label{eq:defdelta}
\Delta \x^\star = \argmin_{\d}  (\v^\star)^\top \d + \frac{1}{2} 
\d^\top \H_{\star} \d + h_1(\x^\star + \d) 
\end{equation}
and
$\v^\star =  \nabla f_1(\x^\star) - \z_{f_2}^\star - \z_{h_2}^\star$, 
$\z_{f_2}^\star = \nabla f_2(\x^\star)$ and $\z_{h_2}^\star \in \partial h_2(\x^\star)$.
\end{lemma}

\textit{Proof~:} Let us start by characterizing the solution $\Delta
\x^\star$.  By definition, we have $\Delta \x^\star + \x^\star = \prox_{h_1}^{\H_\star}(
\x^\star- \H_{\star}^{-1}\v^\star)$ and thus according to the optimality
condition of the proximal operator, the following equation holds
$$
\H_\star( \x^\star - \H_{\star}^{-1}\v^\star - \Delta \x^\star - \x^\star)
\in \partial h_1(\Delta \x^\star + \x^\star)
$$
which after rearrangement is equivalent to
\begin{equation}\label{eq:conditionprox}
\z_{h_2}^\star - \H_\star \Delta \x^\star 
\in \cor{\nabla f(\x^\star)} + \partial h_1(\Delta \x^\star + \x^\star) 
\end{equation}
\GGCorr{}{with $\nabla f(\x^\star) = \nabla f_1(\x^\star) - \nabla f_2(\x^\star)$.}
This also means that there exists a $\z_{h_1\Delta}^\star \in \partial h_1(\Delta \x^\star + \x^\star) $ so that 
\begin{equation}\label{eq:conditionprox2}
\z_{h_2}^\star - \H_\star \Delta \x^\star 
- \cor{\nabla f(\x^\star)}
-\z_{h_1\Delta}^\star = \0
\end{equation}
Remember that by hypothesis, since $\x^\star$ is a stationary point of  Problem (\ref{eq:optprob}), we have
$$
\0 \in\cor{\nabla f(\x^\star)}+ \partial h_1(\x^\star) - \partial h_2(\x^\star)
$$

We now prove that if $\x^\star$ is a stationary point
of  Problem (\ref{eq:optprob}) then $\Delta\x{^*} = \0$ by showing the
contrapositive.
Suppose that $\Delta \x^\star \neq \0 $.  $\Delta \x^\star $ is a vector that
satisfies the optimality condition (\ref{eq:conditionprox})
and it is the unique one according to properties of the
proximal operator. This means that 
the vector $\0$ is not optimal for the problem (\ref{eq:defdelta}) and
thus it does not exist a vector 
\GGCorr{a}{} $\z_{h_1\0}^\star \in \partial h_1(\d + \x^\star) $ so that 
\begin{equation}
\z_{h_2}^\star - \H_\star \d 
- \cor{\nabla f(\x^\star)}
-\z_{h_1\0}^\star = \0
\end{equation}
with $\d=\0$. Note that this equation is valid for any $\z_{h_2}^\star$
chosen in the set $\partial h_2(\x^\star)$ and the above
equation also translates in $\not\exists, \z_{h_1\0}^\star \in \partial h_1( \x^\star) $ so that $
\nabla f(\x^\star)+
\z_{h_1\0}^\star - \z_{h_2}^\star = \0$, which proves that
$\x^\star$ is not a stationary point of problem (\ref{eq:optprob}).

Suppose now that $\Delta \x^\star = \mathbf{0}$, then according to the
definition of $\Delta \x^\star$ and the resulting condition
(\ref{eq:conditionprox}), it is straightforward to note that
$\x^\star$ satisfies the definition of a stationary point. 
\hfill$\square$\\

Now, we proceed by showing that at each iteration, the
search direction $\Delta \x_{k}$ satisfies a property which
implies that for a sufficiently small \GGCorr{step}{} step size $t_k$, the search direction
is a descent direction.
\begin{lemma}
  For $\x_k$ in the domain of $f$ and supposing
that $\H_k \succ 0$ then $\Delta \x_k $ is so that
\begin{align*}
F(\x_{k+1}) &\leq F(\x_k) + t_k \Big( \v_k ^\top
\Delta \x_k + h_1(\Delta \x_k +  \x_k) - h_1(\x_k) \Big)
 \\&~+
O(t_\cor{k}^2)
\end{align*}
and 
\begin{equation} \label{eq:descent}
F(\x_{k+1}) - F(\x_k) \leq  -  t_k \Delta \x_k^\top \H_k \Delta
\x_k + O(t_k^2)
\end{equation}
with $\v_k =  \nabla f_1(\x_k) - \z_{f_2} - \z_{h_2}$.
\end{lemma}
\textit{Proof:} For a sake of clarity, we have dropped the index $k$ and  used
the following notation. $\x := \x_k$, $\Delta \x := \Delta \x_k$, $\x_+:= \x_k + t_k \Delta \x_k$. By definition, we have 
\begin{align*}
  F(\x_{+}) - F(\x) &= f_1(\x_+) -f_\cor{1}(\x) - f_2(\x_+) +
f_2(\x) \\\nonumber 
& + h_1(\x_+) - h_1(\x) -h_2(\x_+) + h_2(\x)\GGCorr{}{.} \nonumber
\end{align*}
\GGCorr{then}{Then} by convexity of $f_2$, $h_2$, $h_1$ and \cor{for $t \in [0, 1]$}, we respectively
have
$$
-\z_{f_2}^\top (\x_+ - \x) \geq f_2(\x) - f_2(\x_+),
$$
$$
-\z_{h_2}^\top (\x_+ - \x) \geq h_2(\x) - h_2(\x_+)
$$
and
$$
h_1(\x + t\Delta \x) \leq t h_1(\x + \Delta
\x) + (1-t) h_1(\x)
$$
Plugging these inequalities in the definition of   $F(\x_{+}) - F(\x)$
gives~:
\begin{align}
\label{eq:desc_eq_intermediaire}
  F(\x_{+}) - F(\x) &\leq f_1(\x_+) -f_1(\x) + (1-t) h_1(\x) \\\nonumber
& \;\;+ t h_1(\x + \Delta \x) \\\nonumber &\;\;{- t(\z_{f_2} + \z_{h_2})^\top\Delta \x 
- h_1(\x) }\\\nonumber
& \leq t \nabla f_1(\x)^\top \Delta \x  + t h_1(\x + \Delta \x) 
\\ &\;\; -t h_1(\x) - t(\z_{f_2} + \z_{h_2})^\top\Delta \x + O(t^2)\nonumber
\end{align}
which proves the first inequality of the lemma.  

For showing the descent property, we demonstrate that
the following inequality holds 
\begin{equation} \label{eq:descendineq}
\underbrace{\v^\top \Delta \x + h_1(\x + \Delta \x) - h_1(\x)}_D
\leq - \Delta \x^\top \H \Delta \x
\end{equation}
Since $\Delta \x$ is the minimizer of Problem (\ref{eq:search}), the
following equation holds \GGCorr{}{for $t \Delta \x$ and $t \in [0, 1]$:}
\begin{align}
&\frac{1}{2} \Delta \x^\top \H\Delta \x  +h_1(\x +
\Delta \x)  + {\v}^\top\Delta \x \\\nonumber &\leq
\frac{t^2}{2} \Delta \x^\top \H\Delta \x  + h_1(\x + t
\Delta \x)  + t {\v}^\top\Delta \x \\\nonumber
&\leq 
\frac{t^2}{2} \Delta \x^\top \H\Delta \x  + (1-t) h_1(\x)
+ t h_1(\x + 
\Delta \x) \\\nonumber &\quad { + t {\v}^\top\Delta \x }
\end{align}
After rearrangement we have the inequality
$$
\v^\top \Delta \x + h_1(\x + \Delta \x) - h_1(\x)
\leq -\frac{1}{2}(1+t)\Delta \x^\top \H \Delta \x
$$
which is valid for all $t \in [0,1]$ and in particular
for $t=1$ which concludes the proof of inequality.  By plugging this result into inequality \GGCorr{(\ref{eq:descendineq})}{(\ref{eq:desc_eq_intermediaire})},  the descent property holds.\hfill $\square$\\

Note that the descent property is supposed to hold for sufficiently small
step size. In our algorithm, this stepsize $t_k$ is selected by backtracking
so that the following sufficient descent condition holds
\begin{equation} \label{eq:descentcond}
  F(\x_{k+1}) - F(\x_k) \leq \alpha t_k D_k
\end{equation}
with $\alpha \in (0,1/2)$. The next lemma shows that if the function $f_1$ is sufficiently smooth,
then there always exists a step size so that the above sufficient descent
condition holds.

\begin{lemma}\label{lem:step}
For $\x$ in the domain of $f$ and assuming that $\H_k \succeq m \I$ with
$m >0$ and $\nabla f_1$ is  Lipschitz with constant $L$ then the
sufficient descent condition in Equation (\ref{eq:descentcond}) holds for all
$t_k$ so that
$$
t_k \leq \min \left(1, 2m \frac{1 - \alpha}{L} \right) 
$$   
\end{lemma}
\textit{Proof~:} This technical proof has been post-poned to
the \GGCorr{supplementary material}{appendix}.\hfill $\square$\\

According to the above lemma, we can suppose that if some
mild conditions on $f_1$ are satisfied (smoothness and bounded curvature)
then, we can expect our DC algorithm to behave properly. This intuition is formalized in the following property.\\
\begin{proposition}\label{prop:sequence}
  Suppose $f_1$ has a gradient which is Lipschitz continuous with constant
$L$ and that $\H_k \succeq m \I$ for all $k$ and $m>0$, then  
all the limit points of 
the sequence $\{\x_k\}$  are stationary points.
\end{proposition}
 \textit{Proof~:}
Let $\x^\star$ be a limit point of the sequence $\{\x_k\}$ then, there
exists a subsequence $\mathcal{K}$ so that
$$
\lim_{k \rightarrow \mathcal{K}} \x_k = \x^\star
$$

At each iteration the step size $t_k$ has been chosen so as 
to satisfy the sufficient descent condition given in Equation 
(\ref{eq:descentcond}). 
According to the above Lemma \ref{lem:step}, the step size $t_k$ is
chosen so as to ensure a sufficient descent and we \GGCorr{now}{know} that 
such a step size always exists and
it is always non-zero. Hence the sequence $\{F(\x_k)\}$ is a strictly decreasing sequence. As $F$ is lower bounded, the sequence $\{F(\x_k)\}$ converges
to some limit. Thus, we have
$$
\lim_{k \rightarrow \infty} F(\x_k) = \lim_{k \rightarrow \mathcal{K}} F(\x_k) = F(\x^\star)
$$
as $F(\cdot)$ is continuous.
Thus, we also have 
$$\lim_{k \rightarrow \mathcal{K}} F(\x_{k+1}) - F(\x_{k}) =  0$$ Now
because each term  $F(\x_{k+1}) - F(\x_{k}\GGCorr{}{)}$ is negative, we can also deduce from Equations (\ref{eq:descendineq}) and (\ref{eq:descentcond}) 
and the limit of $F(\x_{k+1}) - F(\x_{k})$ that
  $$
\lim_{k \rightarrow \mathcal{K}} \v_k^\top \Delta \x_k + h_1(\x_k + \Delta \x_k) - h_1(\x_k) = \lim_{k \rightarrow \mathcal{K}} -\Delta \x_k^\top \H_k \Delta \x_k =0
$$
Since {$\H_k$} is positive definite, this also means that
 $$ \lim_{k \rightarrow \mathcal{K}} \Delta \x_k = 0 $$ 
Considering now that $\Delta \x_k$ is a minimizer of Problem (\ref{eq:search}),
we have
$$
\boldsymbol{0} \in \H_k \Delta \x_k + \partial h_1(\x_k + \Delta \x_k) + \nabla f_1(\x_k)
- \nabla f_2(\x_k) - \partial h_2(\x_k)
$$
Now, by taking limits on both side of the above equation for $k \in \mathcal{K}$,
we have
$$
\boldsymbol{0} \in \partial h_1(\x^\star) + \nabla f_1(\x^\star) - \partial h_2(\x^\star) - \nabla f_2(\x^\star) 
$$
Thus, $\x^\star$ is a stationary point of Problem (\ref{eq:optprob}).\hfill $\square$\\

\rrev{The above proposition shows that under simple conditions on
$f_1$, any limit point of the sequence $\{\x_k\}$ is a stationary point
of $F$. Hence the proposition is quite general and applies to a large class
of functions. If we impose stronger constraints on the functions
$f_1$, $f_2$, $h_1$ and $h_2$, it is possible to leverage on the technique of 
Kurdyka-Lojasiewicz (KL) theory \cite{attouch2010proximal}, recently developed for the convergence
analysis of iterative algorithms for non-convex optimization,
for showing that the sequence $\{\x_k\}$ is indeed convergent.
Based on the recent works developed in Attouch et al. \cite{attouch2010proximal,attouch2013convergence}, Bolte et al. \cite{bolte2013proximal} and
Chouzenoux et al. \cite{chouzenoux13:_variab_metric_forwar_backw},
we have carried out a convergence analysis of our algorithm for
functions $F$ that satisfies the  KL property. However, due to the strong restrictions imposed
by the convergence conditions (for instance on the loss function and
on the regularizer) and for a sake of clarity, we have post-poned
such an analysis to the appendix.}

\section{Experiments}
\label{sec:expe}
In order to provide evidence on the benefits of the \cor{proposed} approach
for solving DC non-convex problems, we have 
carried out two numerical experiments. 
First we analyze our algorithm when the function
$f$ is convex and the regularizer $h$ is a non-convex {and a non-differentiable} sparsity-inducing penalty. Second, we study the case when both $f$ and $h$
are non-convex. 
{All experiments have been run on a Notebook Linux machine powered
by a Intel Core i7 with 16 gigabytes of memory. All the codes have been
written in Matlab.}

{Note that for all numerical results, we have used a limited-memory
  BFGS (L-BFGS) approach for approximating the Hessian matrix $\H_k$
  through rank-1 update. This approach is well known for its ability to handle
  large-scale problems. By default, the limited-memory size for
the L-BFGS has been set to $5$.  
}

\subsection{Sparse Logistic Regression}
We consider here $f(\x)$ as the following convex loss function
$$
f(\x)=\sum_{i=1}^\ell \log(1+ \exp (- y_i \a_i^\top \x) )
$$
where $\{\a_i,y_i\}_{i=1}^\ell$ are the training examples \cor{and their
associated labels} available for learning the model. The regularizer we have
considered is the capped-$\ell_1$ defined as
$h(\x)= h_1(\x) - h_2(\x)$  with 
\begin{equation}
h_1(\x)= \lambda \|\x\|_1   \text{ and }
h_2(\x)= \lambda \big(\|\x\|_1 - \theta \big)_+\label{eq:regterm}
\end{equation}
and the operator $(u)_+ = u$ if $u \geq 0$  
and $0$ otherwise. {Note that here we focus on binary classification
problems but extension to multiclass problems can be easily handled
by using a multinomial logistic loss instead of a logistic one.}

Since several other algorithms are able to solve the optimization
problem related to this sparse logistic regression problem as given by
Equation (\ref{eq:optprob}), our objective here is to show that the
{proposed} DC proximal Newton is computationally more efficient than
competitors, while achieving equivalent classification
performances. For this experiment, we have considered as 
a baseline, a DCA algorithm \cite{thi05:_dc_differ_convex_funct_progr} and single
competitor which is the recently proposed GIST
algorithm\cite{gong2013jieping}. Indeed, this latter approach has
already been shown by the authors to be more efficient than several
other competitors including SCP \GGCorr{}{(sequential convex
  programming)} \cite{lu12:_sequen_convex_progr_method_class},
MultiStage Sparsa
\cite{zhang10:_analy_multi_convex_relax_spars_regul}. 
\modif{As shown in Table \ref{tab:listmethods}, none of these competitors
\GGCorr{can}{} handle second-order information for a non-convex regularization
term. But the computational advantage brought by using this second order information has still to be shown since in practice, the resulting numerical cost per iteration is more important in our approach because of the metric term $\H_k$. As second-order methods usually suffer more for high-dimensionality problems, the comparison
has been carried out when the dimensionality $d$ is very large.
}
Finally,  a
slight advantage has been provided to GIST as we consider its
non-monotone version (more efficient than the monotone counterpart)
whereas our approach decreases the objective value at each iteration.
  {Although DC algorithm  as described in section {\ref{subsec:related_works}} has
already  been
shown to be less efficient than GIST in \cite{boisbunon2014active},
we have still reported its results in order to confirm this tendency. Note that for the DC approach,
we allowed a maximum of $20$ DC iterations.
}

\subsubsection{Toy dataset}

We have firstly evaluated the baseline DC algorithm, GIST and our DC proximal Newton on a toy dataset where only few features are relevant for the discrimination
task.
The toy problem is the same as the one used by \citet{rakotomamonjy11:_ell_p_ell_q_penal}.  The task is a
binary classification problem in $\R^d$.  Among these $d$ variables,
only $T$ of them define a subspace of $\R^d$ in which classes can be
discriminated.  For these $T$ relevant variables, the two classes
follow a Gaussian pdf with means respectively $\mu$ and $-\mu$ and
covariance matrices randomly drawn from a Wishart distribution.  $\mu$
has been randomly drawn from {$\{-1,+1\}^T$. The other $d-T$}
non-relevant variables follow an \emph{i.i.d} Gaussian probability
distribution with zero mean and unit variance for both classes.  We have respectively sampled $N$, and
$n_{t} = 5000$ number of examples for training and testing. Before learning, the training set has been normalized to zero mean and unit variance and  test set has been rescaled accordingly. \rrev{The hyperparameters
$\lambda$ and $\theta$ of the regularization term \eqref{eq:regterm} have been
roughly set so as
to maximize the performance of the GIST algorithm on the test set}.
We have chosen to initialize  all algorithms \cor{with} zero vector
($\x_0=\boldsymbol{0}$) and we 
terminate them  if the relative change of 
two consecutive objective function values is less than
$10^{-6}$. 

Reported performances and running times averaged over $30$ trials are depicted in Table~\ref{tab:toy} for two different settings of the dimensionality
$d$ and the number of training examples $N$.
\begin{table*}[t]
  \centering
\caption{Comparison {between DCA, GIST} and our DC Proximal Newton on toy problems with increasing number of relevant variables. 
Performances reported in \textbf{bold} are statistically significantly different than their competitor counterpart according to
a Wilcoxon signed rank test with a p-value at $0.001$. A minus sign in the relative objective value indicates
that the DC Proximal Newton approach provides larger objective value than GIST. The hyperparameters
$\lambda$ and $\theta$ have been chosen so as to maximize performances of GIST.}
  \label{tab:toy}

\small{
  \begin{tabular}[h]{r|ccc|ccc|cc}

\multicolumn{9}{c}{d= 2000,  N= 100000,  $\lambda=2.00$  $\theta=0.20$} \\\hline
&\multicolumn{3}{c}{Class. Rate (\%)} & \multicolumn{3}{c}{Time (s)} & \multicolumn{2}{c}{ Obj Val (\%)} \\
 T & DCA & GIST & DC-PN  & DCA &  GIST & DC-PN
 &  \multicolumn{2}{c}{ Rel. Diff} \\ \hline

 50 & 92.18$\pm$0.0 & 92.18$\pm$0.0 & 91.94$\pm$0.0 &   255.40$\pm$0.0 &  95.42$\pm$0.0  & 70.17$\pm$0.0 & -6.646 \\
 100 & 91.84$\pm$1.9 & 91.84$\pm$1.9 & 91.78$\pm$1.9 &   117.07$\pm$21.4 &  60.02$\pm$9.9  & \textbf{44.42$\pm$12.0} & -1.095 \\
 500 & 91.52$\pm$0.8 & 91.52$\pm$0.8 & 91.50$\pm$0.8 &   137.85$\pm$14.1 &  57.41$\pm$5.2  & 46.87$\pm$13.0 & -0.339 \\
 1000 & 91.69$\pm$0.7 & 91.69$\pm$0.7 & 91.69$\pm$0.7 &   148.97$\pm$9.9 &  61.18$\pm$6.4  & \textbf{49.05$\pm$15.6} & -0.198 \\\hline
     \multicolumn{9}{c}{} \\
\multicolumn{9}{c}{} \\
\multicolumn{9}{c}{d= 10000,  N= 5000,  $\lambda=2.00$  $\theta=2.00$} \\\hline
&\multicolumn{3}{c}{Class. Rate (\%)} & \multicolumn{3}{c}{Time (s)} & \multicolumn{2}{c}{ Obj Val (\%)} \\
 T & DCA & GIST & DC-PN  & DCA &  GIST & DC-PN
 &  \multicolumn{2}{c}{ Rel. Diff} \\ \hline
 50 & 88.55$\pm$2.5 & 88.53$\pm$2.5 & 88.57$\pm$2.5 &   96.28$\pm$30.4 &  48.82$\pm$11.5  & \textbf{26.54$\pm$2.3} & 0.025 \\
 100 & 87.81$\pm$2.8 & 87.76$\pm$2.8 & 87.81$\pm$2.8 &   72.55$\pm$7.6 &  38.30$\pm$6.6  & \textbf{24.27$\pm$2.5} & 0.016 \\
 500 & 81.82$\pm$0.9 & 81.78$\pm$0.9 & 81.82$\pm$0.9 &   71.91$\pm$6.0 &  33.73$\pm$2.7  & \textbf{21.67$\pm$0.9} & 0.004 \\
 1000 & 76.23$\pm$0.9 & 76.20$\pm$0.9 & 76.23$\pm$0.9 &   74.41$\pm$7.9 &  32.79$\pm$3.2  & \textbf{21.59$\pm$0.9} & 0.007 \\
\hline

 \end{tabular}}

\end{table*}
\rrev{We note that for both problems our DC proximal Newton 
is computationally more efficient
than GIST, with respect to the stopping criterion we set,  while the recognition performances of both approaches  are equivalent.}
{As expected and as discussed 
above, the DC algorithm is substantially
slower than GIST and our approach.}
\modif{Interestingly, we can remark that the competing algorithms
do not reach  similar  objective values. This means that despite having
the same
initialization to the null vector,  all methods
have a different trajectories during optimization and
converge to a different stationary point. Although
we leave the full understanding of this phenomenon to future
works, we conjecture that this is due to the primal-dual
nature of the DC algorithm \cite{dinh98:_dc_optim_algor_solvin_trust_region_subpr} which is in contrast to the first-order primal descent of GIST.
}

\subsubsection{Benchmark datasets}

The same experiments  have been carried out on real-world high-dimensional learning problems. These datasets are those already used by \cite{gong2013jieping} for illustrating the behaviour of their GIST algorithm. Here, the available examples are split in a training and testing set with a ratio of $80\%-20\%$ and \rrev{hyperparameters have been roughly set to maximize performance of GIST}.

From Table \ref{tab:real}, we can note that while almost equivalent, recognition
performances are sometimes statistically better for one method than
the other although there is no clear winner. \rrev{From the running time point of view, our DC proximal Newton {always exhibits a better behaviour
than GIST}. 
Indeed, its running time is always better, regardless of the dataset, and the difference in efficiency is statistically significantly better for $4$ out of $5$ datasets. }
In addition, we can note that in some situations, the gain in running time
reaches an order of magnitude, clearly showing the benefit of a proximal Newton
approach.  {Note that the baseline DC approach is slower than our DC proximal Newton except for one dataset
where it converges faster than all methods. For this dataset, the DC algorithm needed only very few DC iterations
explaining its fast convergence.}

\begin{table*}[t]
  \centering
\caption{Comparison {between DCA, GIST} and our DC Proximal Newton on real-world benchmark problems. The first columns of the table provide the
name of the datasets, their statistics. Performances reported in \textbf{bold} are statistically significantly different than their competitor counterpart according to
a Wilcoxon signed rank test with a p-value at $0.001$. 
A minus sign in the relative objective value indicates
that the DC Proximal Newton approach provides larger objective value than GIST.  }
  \label{tab:real}

\small{
  \begin{tabular}[h]{ccc|ccc|ccc|cc}
\hline
 &&&\multicolumn{3}{c}{Class. Rate (\%)} & \multicolumn{3}{c}{Time (s)} & \multicolumn{2}{c}{ Obj Val (\%)} \\
 dataset & N & d   & DCA &GIST & DC-PN & DCA   & GIST & DC-PN  &  \multicolumn{2}{c}{ Rel. Diff}\\\hline

 la2  & 2460 &  31472 & 91.32$\pm$0.9 & 91.67$\pm$0.9 & 91.81$\pm$0.9 &   36.61$\pm$11.5 &  45.86$\pm$26.4  &\textbf{ 21.74$\pm$11.9} & -165.544 \\
 sports  & 6864 &  14870 & 97.86$\pm$0.4 & 97.94$\pm$0.3 & 97.94$\pm$0.3 &   88.99$\pm$70.8 &  161.45$\pm$162.6  & \textbf{23.76$\pm$13.7 }& -95.215 \\
 classic  & 5675 &  41681 & 96.93$\pm$0.6 & 97.33$\pm$0.5 & 97.38$\pm$0.5 &  \textbf{3.44$\pm$3.8} &  31.60$\pm$11.7  & 17.44$\pm$7.6 & -418.789 \\
 ohscal  & 8929 &  11465 & 87.05$\pm$0.6 & 87.99$\pm$0.6 & \textbf{89.27$\pm$0.6} &   320.39$\pm$134.5 &  44.78$\pm$21.6  & \textbf{19.13$\pm$25.1} & -85.724 \\
 real-sim  & 57847 &  20958 & 95.16$\pm$0.3 & \textbf{96.28$\pm$0.2} & 96.05$\pm$0.2 &   63.81$\pm$96.3 &  382.70$\pm$813.1  &\textbf{ 23.14$\pm$9.3} & -105.902 \\
\hline

 \end{tabular}}

\end{table*}

\subsection{Sparse Transductive Logistic Regression}

\begin{figure*}[!t]
\centering
~\hfill
\includegraphics[width=5cm]{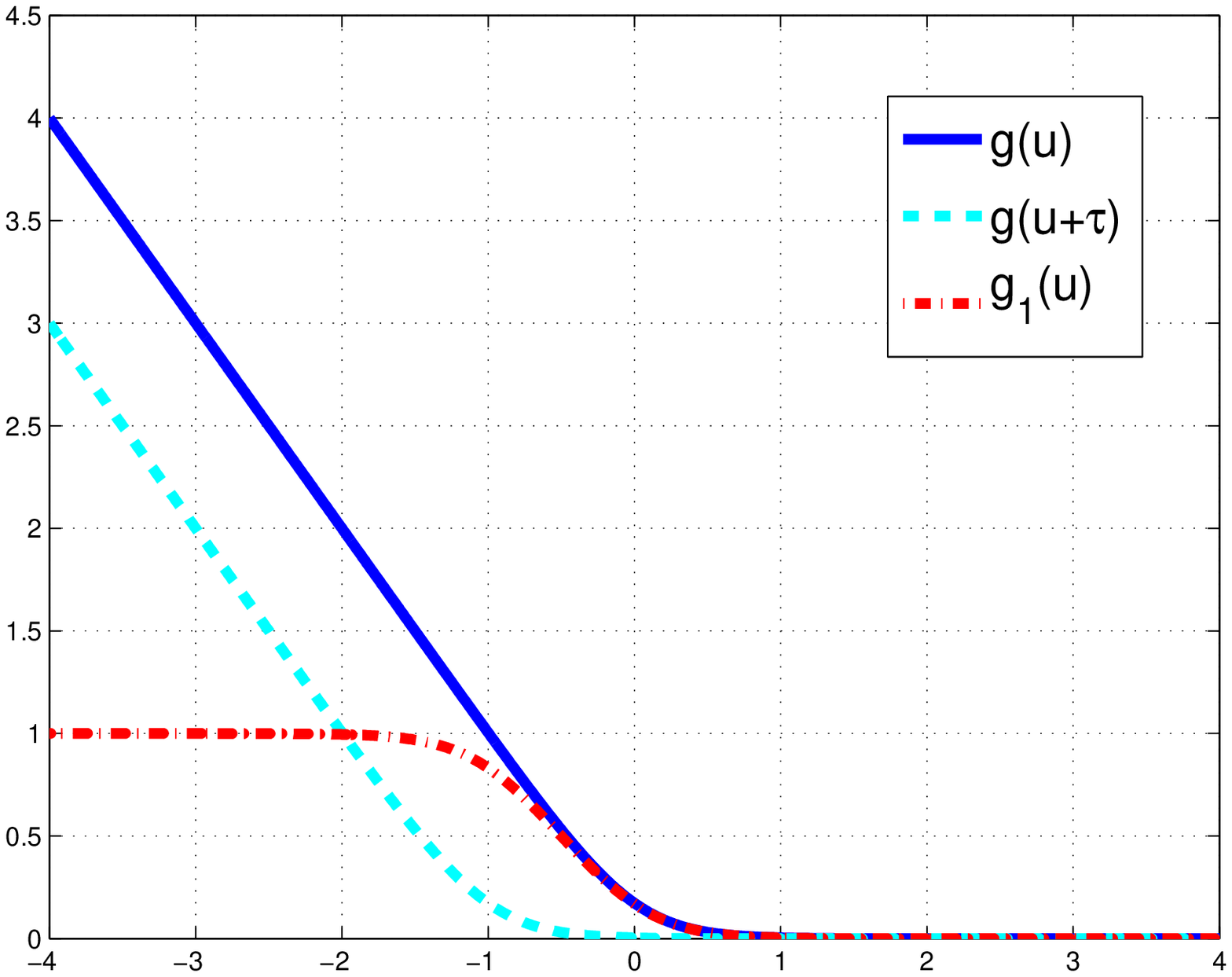}  
~\hfill~
\includegraphics[width=5cm]{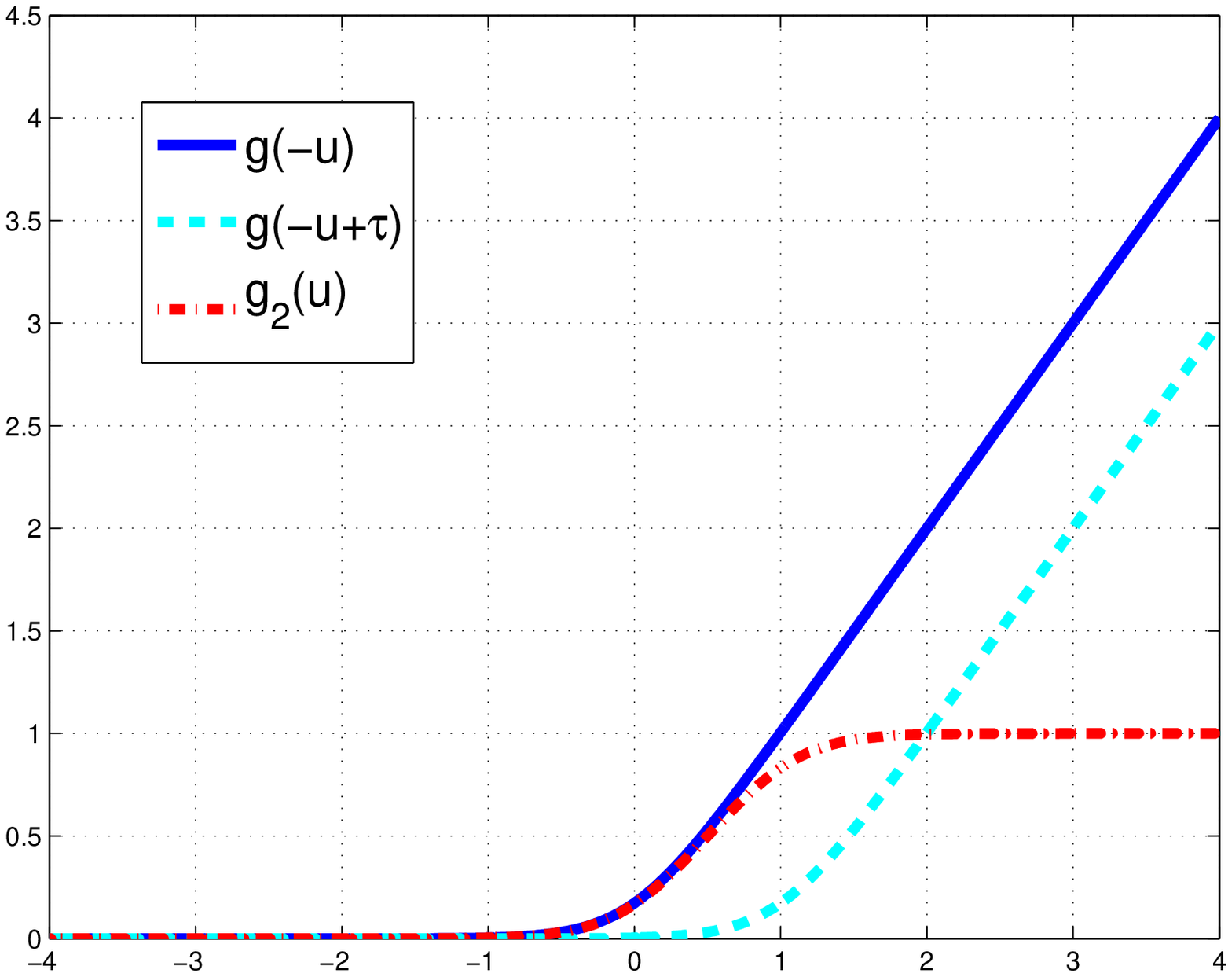}  
~\hfill~
\includegraphics[width=5cm]{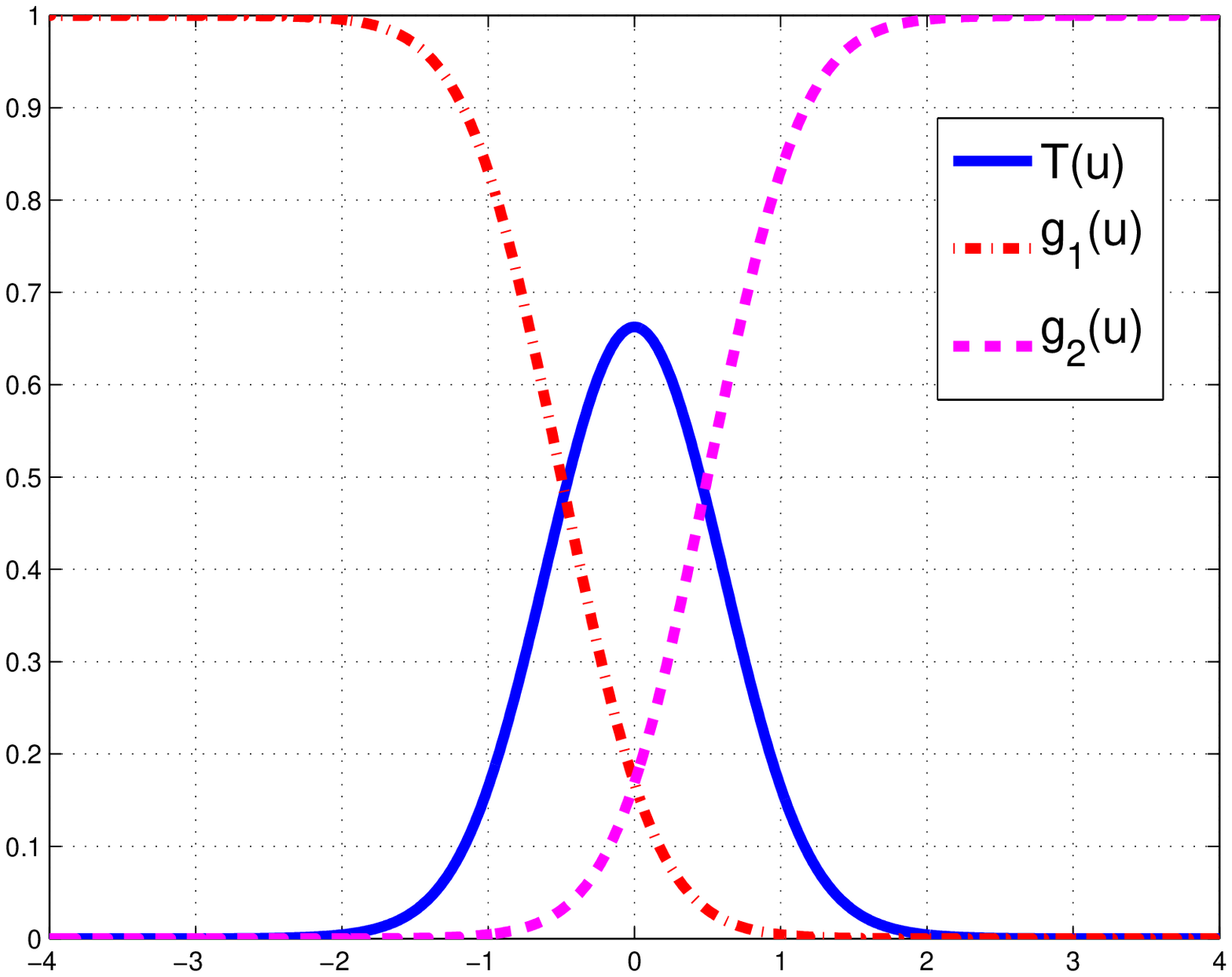}  
~\hfill~
\caption{Example of a non-convex smooth transductive loss function
 {$T(\cdot)$} obtained with $\tau=1$ as well as its components.
(left) $g_1(u)$, (middle) $g_2(u)$, (right)  DC decomposition of $T(u)$.}
\label{fig:transdloss}
\end{figure*}

In this other experiment, we show an example of situation where one
has to deal with a non-convex loss function as well as a non-convex regularizer, namely~: sparse transductive logistic regression. 
{The principle of transductive learning is to leverage unlabeled
  examples during the training step. This is usually done by using a
  loss function for unlabeled examples  that enforces the
decision function  to lie in regions of low density.
A way to achieve this is the use of a symmetric loss function which
penalizes unlabeled examples lying in the margin of the
  classifier. It is well known that this approach, also known as low
  density separation, leads to non-convex data fitting term on the
  unlabeled examples \cite{chapelle05:_semi}}.
For instance, {Joachims  \cite{joachims_transd}} has considered
a Symmetric Hinge loss for the unlabeled examples in their
transductive implementation of SVM. 
Collobert et al. \cite{tsvm_collobert_jmlr_2006} extended this
idea of symmetric Hinge loss into a symmetric ramp loss, which has
a plateau on its top.   
In order
to have a smooth transductive loss, Chapelle et al \cite{chapelle05:_semi}  used a symmetric sigmoid loss.  

{
For our purpose the transductive loss function is required to be differentiable. Hence we propose the following {symmetric} differentiable loss that can be written as a difference of convex function
$$
T(u) = 1- g_1(u) - g_2(u)
$$
 where  $g_1(u)= \frac{1}{\tau} (g(u)-g(u+\tau))$, $g_2(u)=g_1(-u)$ and
$g(u)= \log(1+ \exp(-u))$. Note that $g(u)$ is a convex function as
depicted in Figure \ref{fig:transdloss} and combinations of
shifted and reversed versions of $g(u)$ lead to $g_1$ and $g_2$. 
 $\tau$ is a parameter that modifies the smoothness of $T(\cdot)$. From the expression of $g_1$ and
$g_2$, it is easy to retrieve the difference of convex functions form of $T(u) = T_1(u) - T_2(u)$ with $T_1(u) = 1 + \frac{1}{\tau}\left( g(u+\tau)  + g(-u+\tau)\right)$ and $T_2(u) = \frac{1}{\tau}\left( g(u)  + g(-u)\right)$. The transductive loss {$T(\cdot)$} as well as $g_1$ and $g_2$
and their components  are illustrated in Figure \ref{fig:transdloss}.
}

According to this definition of the transductive loss,  for our experiments, we have used the following loss involving all training examples
\begin{equation}
  \label{eq:transductiveloss}
  \cor{f(\x)=}\sum_{i=1}^\ell \log(1+ \exp (- y_i \a_i^\top \x) )+ \gamma \sum_{j=1}^{\ell_u} T(\b_j^\top \x)
\end{equation}
 $\{\a_i,y_i\}$ being the labeled examples and 
$\{\GGCorr{b}{\b}_j\}$ the unlabeled ones
and $\gamma$ is an hyperparameter that balances the weight of
both \GGCorr{the}{}losses. {As previously the capped-$\ell_1$ serves as a regularizer.}

\subsubsection{Toy dataset}

In order to illustrate the benefit of our sparse transductive approach,
we have considered the same toy dataset as in the previous subsection
and the same
experimental protocol. However, we have considered only $5$ relevant variables, sampled  $100$ training examples and $5000$ testing examples. In addition, we have considered $10000$ unlabeled examples. The total number of  variables
is varying. 
We have compared the recognition performance of $3$ algorithms~:
the above-described capped-$\ell_1$ sparse logistic regression, the
non-sparse transductive SVM (TSVM) of \citet{chapelle05:_semi}\footnote{we used
the code available on the author's website.} 
and our sparse transductive logistic regression.

\begin{figure}[t]
\centering
\includegraphics[width=7cm]{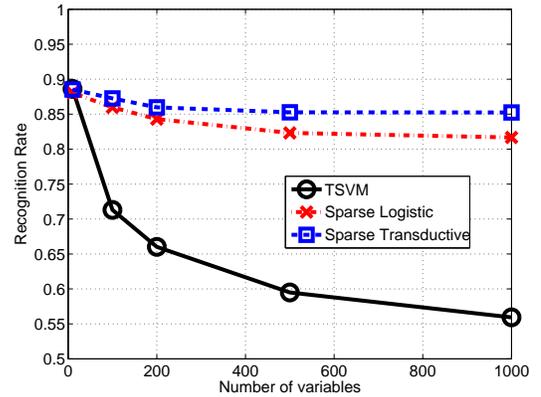}  
\caption{Recognition rate of different algorithms that
are either sparse, transductive or both with respects
to the number of variables in the problem, the number of
relevant variables being $5$.  }
\label{fig:toytransd}
\end{figure}

Evolution of the recognition rate of these algorithms with respects
to the number of variables in the learning problem is depicted
in Figure ~\ref{fig:toytransd}. \rrev{Interestingly,  when the number
of variables is small enough, all algorithms perform equivalently.
Then, as the number of (noisy) variables increases, the transductive
SVM suffers a drop of performances. It seems more beneficial in this
case to consider a model that is able to select relevant 
variables as our  capped-$\ell_1$ sparse logistic regression still
performs good. Best performances are obtained using our DC formulation
introduced for solving the sparse
transductive logistic regression problem which is able to remove noisy
variables and take advantage of the unlabeled examples.}

\subsubsection{Benchmark datasets}

We have also analyzed the benefit of using unlabeled examples in
high-dimensional learning problems. \rrev{For this experiment, all the
hyperparameters of all models have been cross-validated.
For instance, $\lambda$, $\theta$ (parameters of the capped $\ell_1$) and 
$\gamma$ have been respectively searched among the sets $\{0.2,2\}$,
$\{0.2,2\}$ and $\{0.005,0.001,0.005,0.01\}$. 
}
averaged
results over $10$ trials are reported in Table \ref{tab:transd}. 
Note that the results of the transductive SVM of \citet{chapelle05:_semi}
have not been reported because the provided code was
not able to provide a solution in a reasonable amount of time.
\rrev{Results in Table \ref{tab:transd} show that being able to
handle non-convex loss functions, related to the transductive
loss and non-convex sparsity-inducing regularizers helps in achieving better performances in accuracy.  Again, we can remark that the benefits of unlabeled examples  are compelling especially when few labeled examples
are in play. Differences in performances are indeed statistically significant
for most datasets.
}
 {
 In order to further evaluate the accuracy of the proposed
 method in very high-dimensional setting, we have run  the
 comparison on the  \emph{URL} dataset. This dataset involves
 about $3.10^6$ features and we have learned a decision function
 using only $1000$ training examples and $40000$ unlabelled examples.
 Although difference in performances is not significant, leveraging
on unsupervised examples helps in improving accuracy. Note that for
this problem, the average running times of our DC-based sparse logistic regression
and the DC-based sparse transductive regression are respectively about
$500$ and $700$ seconds. This shows that the proposed approach
allows to handle large-scale and very high-dimensional learning problems. 
 }
\begin{table}[t]
  \centering
\caption{Comparing the recognition rate of a sparse logistic regression and
a sparse transductive logistic regression both with capped-$\ell_1$ regularizer.
$\ell$ and $\ell_u$ respectively denotes the number of \GGCorr{training}{labeled} and unlabeled examples.  }
\label{tab:transd}
  \begin{tabular}[h]{cccccc}
\hline
 \multicolumn{3}{c}{}&\multicolumn{2}{c}{Class. Rate (\%)}   \\
 dataset &$d$& $\ell$ & $\ell_u$   & Sparse Log  & Sparse Transd.  \\\hline
 la2  & 31472& 61 & 2398 &67.65$\pm$2.6 & \textbf{70.23$\pm$3.1 }   \\
 sports  & 14870& 85 & 6778 &81.26$\pm$5.0 & \textbf{88.15$\pm$4.4}     \\
 classic  &41681& 70 & 5604 &72.74$\pm$4.3 & \textbf{86.97$\pm$2.2}     \\
 ohscal  &11465& 55 & 8873 &70.35$\pm$2.4 & \textbf{73.39$\pm$3.6}     \\
 real-sim  &20958& 723 & 57124 &88.81$\pm$0.3 & 88.91$\pm$1.4     \\
 url &3.23$\times 10^6$& 1000 & 40000 &   86.64$\pm$5.8& 87.39$\pm$6.0  \\\hline
\end{tabular}
\end{table}

\section{Conclusions}
\label{sec:conclu}

This paper introduced  a general proximal Newton algorithm that
\GGCorr{optimize}{optimizes} the composite sum of functions. \GGCorr{Its specificity is that
it  is able}{A specificity of the approach is its ability} to deal with the non-convexity of both terms 
while  one of these terms is in addition allowed to be non-differentiable. 
While   most of the works in the machine learning and
optimization communities have been addressing these non-differentiability 
and non-convexity issues separately,
there exists a number of learning problems such as sparse transductive learning that
require efficient optimization scheme on non-convex and non-differentiable
functions. 
Our algorithm is based on two steps: the first one looks for a search
direction through a proximal Newton step while the second
one performs a line search on that direction.  
We also provide in this work the proof that the iterates
generated by this algorithm behaves correctly in the sense
that limit points of the sequences are stationary points. 
 Numerical experiments show that the second order information 
 used in our algorithm through the matrix $\H_k$ allow faster convergence
 than  proximal gradient based descent approaches for non-convex regularizers.
One of the strength of our framework is its ability to handle non-convexity
on both the smooth loss function and the regularizer. We have illustrated
this ability by learning a sparse transductive \GGCorr{logistice}{logistic} regression model.

  For the
 sake of reproducible research, the code source of the numerical
 simulation will be freely available on the authors website.

\section{Appendix}

\subsection{Details on the proximal expression of
$\Delta \x_k$}

We provide in this paragraph the steps for obtaining 
Equation (\ref{eq:searchprox}) from Equation (\ref{eq:search}).

Remind that for a lower semi-continuous convex function $h_1$, the
proximal operator is defined as \cite{combettes11:_proxim_split_method_in_signal_proces} 
$$
\y^\star= \prox_{h_1}^{\H}(\x)= \arg\min_\y \frac{1}{2} \|\y - \x\|_{\H}^2 + h_1(\y)
$$
$\y^\star$ can be characterized by the optimality condition of the optimization
problem which is
$$-\H( \y^\star - \x) \in \partial h_1(\y^\star)$$

The search direction is provided by Equation (\ref{eq:search})
which we remind is
\begin{equation}\nonumber
  \argmin_{\Delta \x} \frac{1}{2} \Delta \x^\top \H_k\Delta \x  +h_1(\x_k +
\Delta \x)  + {\v_k}^\top\Delta \x 
\end{equation}
By posing $\z= \x_k + \Delta \x$, we can equivalently look at a shifted version of this problem:
\begin{equation}\nonumber
  \z_k=\argmin_{\z} \frac{1}{2} (\z- \x_k)^\top \H_k(\z- \x_k)   +h_1(\z)  + \v_k^\top(\z- \x_k)
\end{equation}
Optimality condition of this problem is
$$-\H_k(\z_k- (\x_k - \H_k^{-1}\v_k )) \in \partial h_1(\z_k)$$

Hence, according to the optimality condition of the proximal operator,
we have
$$
\z_k= \prox_{h_1}^{\H}(\x_k - \H_k^{-1}\v_k )
$$
and thus 
$$
\Delta \x_k = \prox_{h_1}^{\H}(\x_k - \H_k^{-1}\v_k ) - \x_k
$$
which is Equation (\ref{eq:searchprox}).

\subsection{Lemma \ref{lem:step} and proof}

\emph{Lemma 3 : }For $\x$ in the domain of $f$ and assuming that $\H_k \succeq m \I$ with
$m >0$ and $\nabla f_1$ is  Lipschitz with constant $L$ then the
sufficient condition in Equation (\ref{eq:descentcond}) holds for all
$t_k$ so that
$$
t_k \leq \min \left(1, 2m \frac{1 - \alpha}{L} \right) 
$$   
\textit{Proof~:} \GGCorr{}{Recall that $\x_+:= \x_k + t_k \Delta \x_k$.}
By definition, we have
\begin{align*}
  F(\x_{+}) - F(\x) &= f_1(\x_+) -f_1(\x) - f_2(\x_+) +
f_2(\x) \\\nonumber 
& + h_1(\x_+) - h_1(\x) -h_2(\x_+) + h_2(\x).
\end{align*}
Then by convexity of $f_2$, $h_2$ and $h_1$, we derive that \GGCorr{}{(see equation (\ref{eq:desc_eq_intermediaire}))}
\begin{align*}
  F(\x_{+}) - F(\x) &\leq f_1(\x_+) -f_1(\x) + (1-t) h_1(\x) \\\nonumber
& \;\;+ t h_1(\x + \Delta \x) - (\z_{f_2} + \z_{h_2})^\top (t\Delta \x) \\\nonumber
& - h_1(\x) \nonumber
\end{align*}
According to a Taylor-Laplace formulation, we have~:
$$
f_1(\x_+) -f_1(\x) = \int_0^1 \nabla f_1(\x +s t \Delta \x)^\top
(t\Delta \x) ds 
$$
thus, we can rewrite
\begin{align} \nonumber
  F(\x_{+}) - F(\x) &\leq \int_0^1 \nabla f_1(\x +s t \Delta \x)^\top
(t\Delta \x) ds
  \GGCorr{+}{-} t h_1(\x) \\\nonumber
& \;\;+ t h_1(\x + \Delta \x) - (\z_{f_2} + \z_{h_2})^\top (t\Delta \x)   \\\nonumber
& \leq \int_0^1 \Big(\nabla f_1(\x +s t \Delta \x) - \nabla f_1(\x)  \Big)^\top
(t\Delta \x)  ds  \\\nonumber
& \;\;+ t h_1(\x + \Delta \x) +\nabla f_1(\x)^\top(t\Delta \x) \\\nonumber
&- (\z_{f_2} + \z_{h_2})^\top (t\Delta \x) 
- t h_1(\x) \nonumber \\\nonumber
&  \leq t \Big( \int_0^1 \Big(\nabla f_1(\x +s t \Delta \x) - \nabla f_1(\x)  \Big)^\top
(\Delta \x)  ds  \\\nonumber
& \;\;+  h_1(\x + \Delta \x) +\nabla f_1(\x)^\top(\Delta \x) \\\nonumber 
& - (\z_{f_2} + \z_{h_2})^\top (\Delta \x) 
-  h_1(\x) \Big) \nonumber
\end{align}
Then using Cauchy-Schwartz inequality and the fact that $f_1$ is gradient
Lipschitz of constant $L$, we have~:
\begin{align}\nonumber
F(\x_{+}) - F(\x) &  \leq t \Big( \int_0^1 stL \|\Delta \x\|_2^2  ds  \\\nonumber
& \;\;+  h_1(\x + \Delta \x) +\nabla f_1(\x)^\top(\Delta \x) \\\nonumber
&- (\z_{f_2} + \z_{h_2})^\top (\Delta \x) 
-  h_1(\x) \Big) \nonumber \\\nonumber
& \leq t \Big( \frac{tL}{2} \|\Delta \x\|_2^2   \\\nonumber
& \;\;+  h_1(\x + \Delta \x) +\nabla f_1(\x)^\top(\Delta \x) - \\\nonumber
& (\z_{f_2} + \z_{h_2})^\top (\Delta \x) 
-  h_1(\x) \Big) \nonumber \\\nonumber
& \leq t \Big( \frac{tL}{2} \|\Delta \x\|_2^2   \\\nonumber
& \;\;+  h_1(\x + \Delta \x) -  h_1(\x) \GGCorr{}{+ \v_k^\top(\Delta \x)  } \Big) 
 \nonumber \\\nonumber
& \leq t \Big( \frac{tL}{2} \|\Delta \x\|_2^2 + D \Big)  
\end{align}
Now, if $t$ is so that
$$
t \leq 2m\frac{1-\alpha}{L}
$$
then 
\begin{align}\nonumber
\frac{Lt}{2} \|\Delta \x\|_2^2 &\leq m (1 - \alpha)
\|\Delta \x\|_2^2 \\\nonumber
& = (1-\alpha) \Delta \x^\top(m \I) \Delta \x \\\nonumber
& \leq (1-\alpha) \Delta \x^\top\H \Delta \x \\\nonumber
&  \leq - (1-\alpha) D
\end{align}
where the last inequality comes from the descent property.
Now, we plug this inequality back and get
$$
t \Big( \frac{tL}{2} \|\Delta \x\|_2^2 + D \Big) \leq
t \Big(  -(1 -\alpha) D + D \Big)
= t \alpha D
$$
which concludes the proof that for all 
$$
t \leq \min \Big (1, 2m\frac{1-\alpha}{L} \Big)
$$
we have 
$$
F(\x_{+}) - F(\x) \leq t \alpha D
$$

\subsection{Convergence property for $F$ satisfying the KL property}

\GGCorr{The general convergence property we have given in Proposition \ref{prop:sequence} is rather weak but it applies to a large class of functions. If
we restrict the classes of functions and impose further conditions on 
the algorithms and some of its parameters, by leveraging on the so-called
Kurdyka-Lojasiewiszc property, it becomes possible to get
stronger convergence property, for instance, we can prove
the convergence of the sequence $\{\x_k\}$ to a stationary point of
$F(\x)$. }{Proposition \ref{prop:sequence} provides the general convergence property of our algorithm that applies to a large class of functions. Stronger convergence property (for instance, the convergence of the sequence $\{\x_k\}$ to a stationary point of $F(\x)$) can be attained  by restricting the class of functions and by imposing further conditions on  the algorithms and some of its parameters. For instance, by considering functions  $F(\x)$ that
satisfy the so-called Kurdyka-Lojasiewiszc property,  convergence of the sequence can therefore be  established.}

\begin{proposition}\label{prop:cvgeKL}
\GGCorr{Let us assume that the following hypotheses }{Assume the following assumptions:}
\begin{itemize}
\item hypotheses on $f$ and $h$ given in section \ref{section:fram}  are
satisfied 
\item $h$ is continuous and defined over $\R^d$
\item  $\H_k$ is so that $\H_k \succeq m \I$ for all $k$ and $m>0$. 
\item  $F$ is coercive and it satisfies the Kurdyka-Lojasiewicz property, 
\item \GGCorr{\remii{$h_2$?} $f_2$}{$h_2$} verifies the
$L_2$-Lipschitz gradient property, and thus  
there exists constant  $L_{h_2}$ 
$$
\| \u - \v \|_2 \leq L_{h_2} \|\x - \y\|_2 \quad   \u \in \partial h_2(\x) \text{ and } \v \in \partial h_2(\y)
$$
\item at each iteration, $\H_k$ is so that the function $\tilde f_1(\z,\x_k)= f_1(\x_k) + \nabla f_1(\x_k)^\top(\z - \x_k)
+ \frac{1}{2} \|\z-\x_k\|^2_{\H_k}$ is a majorant approximation of $f_1(\cdot)$ \emph{i.e}
$$
f_1(\z) \leq \tilde f_1(\z,\x_k) \quad \forall \z
$$
\item there exists an $\tilde\alpha \in (0,1]$ so that at each iteration the condition
$$
F(\x_{k+1}) \leq (1-\tilde \alpha) F(\x_k) + \tilde \alpha F(\z_k)
$$ 
holds. Here, $\z_k $ is equal to $\x_k + \Delta \x$ as defined in Appendix A.
\end{itemize}
Under the above assumptions, the sequence $\{\x_k\}$  generated by our algorithm
(\ref{alg:algo}), converge to a critical point of $F=f+h$. \\
\end{proposition}

{Before stating the proof, let us note that these conditions are quite restrictive and thus it may
limit the scope of the convergence property. For instance, the hypothesis
on $h_2$ holds for the SCAD regularizer but does not hold for
the capped-$\ell_1$ penalty. We thus leave for future works the development
of an adaptation of this proximal Newton algorithm for which
convergence of the sequence $\{\x_k\}$ holds for a larger
class of regularizers and loss functions.

\textit{Proof:} The proof of convergence of sequence $\{\x_k\}$ strongly relies
on Theorem 4.1 in \cite{chouzenoux13:_variab_metric_forwar_backw}. Basically, this theorem
states that sequences $\{\x_k\}$ generated by an algorithm  minimizing a function
$F =f +h$  with $h$ being convex and $F$ satisfying Kurdyka-Lojasiewicz property converges to a stationary point of 
$F$ under the above assumptions. The main difference between our framework and the one \GGCorr{of Chouzenoux et al.}{in \cite{chouzenoux13:_variab_metric_forwar_backw}} is that  we consider a non-convex function
$h(\x)$. Hence, for a sake of brevity, we have given in what follows only some parts of the  proofs \GGCorr{Chouzenoux's}{given in \cite{chouzenoux13:_variab_metric_forwar_backw}} that  needed to be reformulated due to the non-convexity of $h(\x)$.
  \\

\noindent i) \underline{sufficient decrease property}.  This property provides
similar guarantee than Lemma 4.1 in \cite{chouzenoux13:_variab_metric_forwar_backw}.
This property easily derives from
Equations (\ref{eq:descentcond}) and (\ref{eq:descendineq}). Combining these
two equations tells us
that
$$
F(\x_{k+1}) - F(\x_k) \leq - \alpha t_k \Delta \x^\top \H \Delta \x
$$ 
where by definition, we have $\x_{k+1} = \x_k + t_k \Delta \x$. Thus, we get
\begin{align}\nonumber
F(\x_{k+1}) - F(\x_k) &\leq - \frac{\alpha}{ t_k} \|\x_{k+1} - \x_k\|_{\H_k}^2 \\\nonumber
&  \leq- \frac{\alpha m }{ t_k} \|\x_{k+1} - \x_k\|^2_2 \\\nonumber
&  \leq- \alpha m  \|\x_{k+1} - \x_k\|^2_2
\end{align}
which proves that a sufficient decrease occurs at each iteration of our algorithm. In addition, because $\x_{k+1} - \x_k = t_k \Delta \GGCorr{\x_k}{\x} = t_k (\z_k - \x_k)$, we also have
\begin{equation}
  \label{eq:decrease2}
  F(\x_{k+1})  \leq  F(\x_k) - \alpha m \underline{t}^2  \|\z_{k+1} - \x_k\|^2_2
\end{equation}
where $\underline{t}$ is the smallest $t_k$ we may encounter. According to Lemma
\ref{lem:step}, we know that $\underline{t} > 0$.

ii) \underline{convergence of $F(\z_k)$}
remind  that we have defined $\z_k$ as (see appendix A)
$$
\z_k =  \argmin_{\z}  \frac{1}{2} (\z- \x_k)^\top \H_k(\z- \x_k)   +h_1(\z)  + \v_k^\top(\z- \x_k)
$$
which is equivalent, by expanding $\v_k$ and adding constant terms, to 
\begin{align} \nonumber
 \min_{\z} &   \frac{1}{2} (\z- \x_k)^\top \H_k(\z- \x_k)   +f_1(\x_k) + \nabla f_1(\x_k)^\top(\z- \x_k)   \\\nonumber
&- f_2(\x_k) - \nabla f_2(\x_k)^\top(\z- \x_k) \\\nonumber
& - h_2(\x_k) -  \partial h_2(\x_k)^\top(\z- \x_k)  \\\nonumber
& +h_1(\z) 
\end{align}
Note that the terms in the first  line  of this minimization problem
majorize $f_1$ by hypotheses and the terms in the second and third lines
respectively majorizes $-f_2$ and $-h_2$ since they are concave function.
When we denotes as $Q(\z,\x_k)$ the objective function of the above problem, we have
\begin{equation}
  \label{eq:descentz}
  F(\z_k) \leq Q(\z_k,\x_k) < Q(\x_k,\x_k)= F(\x_k)
\end{equation}
where the first inequality holds because $Q(\z,\x_k)$ majorizes $F(\z)$,
the second one holds owing to the minimization. Combining this last
equation with the assumption on $F(\x_{k+1})$, we have
$$
\tilde \alpha^{-1} \big (F(\x_{k+1}) - (1-\tilde \alpha) F(\x_k)\big) \leq F(\z_k) \leq
F(\x_k) 
$$
This last equation allows us to conclude that if $F(\x_k)$ converges
to a real $\xi$, then $F(\z_k)$ converges to $\xi$.

iii) \underline{bounding subgradient at $F(\z_k)$}

A subgradient $\z_F$ of $F$ at a given $\z_k$ is by definition
$$
\z_F=\nabla f_1(\z_k) - \nabla f_2(\z_k)  + \z_{h_1,z_k} - \z_{h_2,z_k}  
$$
where $\z_{h_1,z_k} \in \partial h_1(\z_k)$ and $\z_{h_2,z_k} \in \partial h_2(\z_k)$. Hence, we have
\begin{align}\nonumber
  \|\z_F\|  \leq & \|\nabla f_1(\z_k) - \nabla f_1(\x_k) \| +  \|\nabla f_2(\z_k) - \nabla f_2(\x_k) \| \\\nonumber
& + \|\z_{h_2,z_k} - \z_{h_2}\| \\\nonumber
& + \| \nabla f_1(\x_k) - \nabla f_2(\x_k) +  \z_{h_1,z_k}   - \z_{h_2} \|
\end{align}

In addition, owing to the optimality condition of $\z_k$, the following hold
$$
\H_k (\z_k - \x_k)= \nabla f_1(\x_k) - \nabla f_2(\x_k) +  \z_{h_1,z_k}   - \z_{h_2}
$$
Hence, owing to the Lipschitz gradient hypothesis of $f_1$ and $f_2$ and 
the hypothesis on $h_2$, there exists a constant $\mu>0$ such that
\begin{equation}
  \|\z_F\| \leq \mu \|\z_k - \x_k\|
\end{equation}


\addcontentsline{toc}{section}{References}
\bibliographystyle{IEEEtran}

\begin{thebibliography}{}
\providecommand{\url}[1]{#1}
\csname url@samestyle\endcsname
\providecommand{\newblock}{\relax}
\providecommand{\bibinfo}[2]{#2}
\providecommand{\BIBentrySTDinterwordspacing}{\spaceskip=0pt\relax}
\providecommand{\BIBentryALTinterwordstretchfactor}{4}
\providecommand{\BIBentryALTinterwordspacing}{\spaceskip=\fontdimen2\font plus
\BIBentryALTinterwordstretchfactor\fontdimen3\font minus
  \fontdimen4\font\relax}
\providecommand{\BIBforeignlanguage}[2]{{%
\expandafter\ifx\csname l@#1\endcsname\relax
\typeout{** WARNING: IEEEtran.bst: No hyphenation pattern has been}%
\typeout{** loaded for the language `#1'. Using the pattern for}%
\typeout{** the default language instead.}%
\else
\language=\csname l@#1\endcsname
\fi
#2}}
\providecommand{\BIBdecl}{\relax}
\BIBdecl

\end{thebibliography}


\begin{thebibliography}{10}
\providecommand{\url}[1]{#1}
\csname url@samestyle\endcsname
\providecommand{\newblock}{\relax}
\providecommand{\bibinfo}[2]{#2}
\providecommand{\BIBentrySTDinterwordspacing}{\spaceskip=0pt\relax}
\providecommand{\BIBentryALTinterwordstretchfactor}{4}
\providecommand{\BIBentryALTinterwordspacing}{\spaceskip=\fontdimen2\font plus
\BIBentryALTinterwordstretchfactor\fontdimen3\font minus
  \fontdimen4\font\relax}
\providecommand{\BIBforeignlanguage}[2]{{%
\expandafter\ifx\csname l@#1\endcsname\relax
\typeout{** WARNING: IEEEtran.bst: No hyphenation pattern has been}%
\typeout{** loaded for the language `#1'. Using the pattern for}%
\typeout{** the default language instead.}%
\else
\language=\csname l@#1\endcsname
\fi
#2}}
\providecommand{\BIBdecl}{\relax}
\BIBdecl

\bibitem{Tibshrani_Lasso_1996}
R.~Tibshirani, ``Regression shrinkage and selection via the lasso,''
  \emph{Journal of the Royal Statistical Society}, vol.~58, no.~1, pp.
  267--288, 1996.

\bibitem{chen_bpjournal}
S.~Chen, D.~Donoho, and M.~Saunders, ``Atomic decomposition by basis pursuit,''
  \emph{SIAM Journal Scientific Comput.}, vol.~20, no.~1, pp. 33--61, 1999.

\bibitem{li10:_two_condit_equiv_norm_solut}
Y.~Li and S.-I. Amari, ``Two conditions for equivalence of 0-norm solution and
  1-norm solution in sparse representation,'' \emph{Neural Networks, IEEE
  Transactions on}, vol.~21, no.~7, pp. 1189--1196, Jul. 2010.

\bibitem{donoho06:_for}
D.~Donoho, ``For most large underdetermined systems of linear equations, the
  minimal $\ell_1$ solution is also the sparsest solution,''
  \emph{Communication in Pure and Applied Mathematics}, vol.~59, no.~6, pp.
  797--829, 2006.

\bibitem{shevade03:_simpl_and_effic_algor_for}
S.~Shevade and S.~Keerthi, ``A simple and efficient algorithm for gene
  selection using sparse logistic regression,'' \emph{Bioinformatics}, vol.~19,
  no.~17, pp. 2246--2253, 2003.

\bibitem{beck09:_fast_iterat_shrin_thres_algor}
A.~Beck and M.~Teboulle, ``A fast iterative shrinkage-thresholding algorithm
  for linear inverse problems,'' \emph{SIAM Journal on Imaging Sciences},
  vol.~2, no.~1, pp. 183--202, 2009.

\bibitem{yuan2012improved}
G.-X. Yuan, C.-H. Ho, and C.-J. Lin, ``An improved glmnet for l1-regularized
  logistic regression,'' \emph{The Journal of Machine Learning Research},
  vol.~13, pp. 1999--2030, Jun. 2013.

\bibitem{bach11:_convex}
F.~Bach, R.~Jenatton, J.~Mairal, and G.~Obozinski, ``Convex optimization with
  sparsity-inducing norms,'' in \emph{Optimization for Machine Learning},
  S.~Sra, S.~Nowozin, and S.~Wright, Eds.\hskip 1em plus 0.5em minus
  0.4em\relax MIT Press, 2011.

\bibitem{zou_adaptive_lasso_2006}
H.~Zou, ``The adaptive lasso and its oracle properties,'' \emph{Journal of the
  American Statistical Association}, vol. 101, no. 476, pp. 1418--1429, 2006.

\bibitem{Fan_LI_scad_2001}
J.~Fan and R.~Li, ``Variable selection via nonconcave penalized likelihood and
  its oracle properties,'' \emph{Journal of the American Statistical
  Association}, vol.~96, no. 456, pp. 1348--1360, 2001.

\bibitem{KnightFuAsymptotics}
K.~Knight and W.~Fu, ``Asymptotics for lasso-type estimators,'' \emph{Annals of
  Statistics}, vol.~28, no.~5, pp. 1356--1378, 2000.

\bibitem{CandesReweighted2008}
E.~Cand{\`e}s, M.~Wakin, and S.~Boyd, ``Enhancing sparsity by reweighted
  $\ell_1$ minimization,'' \emph{J. Fourier Analysis and Applications},
  vol.~14, no. 5-6, pp. 877--905, 2008.

\bibitem{laporte13:_noncon_regul_featur_selec_rankin}
L.~Laporte, R.~Flamary, S.~Canu, S.~Dejean, and J.~Mothe, ``Nonconvex
  regularizations for feature selection in ranking with sparse svm,''
  \emph{Neural Networks and Learning Systems, IEEE Transactions on}, vol.~25,
  no.~6, pp. 1118--1130, 2014.

\bibitem{gasso09:_recov_spars_signal_with_certain}
G.~Gasso, A.~Rakotomamonjy, and S.~Canu, ``Recovering sparse signals with a
  certain family of non-convex penalties and dc programming,'' \emph{IEEE
  Trans. Signal Processing}, vol.~57, no.~12, pp. 4686--4698, 2009.

\bibitem{combettes11:_proxim_split_method_in_signal_proces}
P.~L. Combettes and J.-C. Pesquet, ``Proximal splitting methods in signal
  processing,'' in \emph{Fixed-Point Algorithms for Inverse Problems in Science
  and Engineering}, H.~H. Bauschke, R.~Burachik, P.~L. Combettes, V.~Elser,
  D.~R. Luke, and H.~Wolkowicz, Eds.\hskip 1em plus 0.5em minus 0.4em\relax
  Springer-Verlag, 2011, pp. 185--212.

\bibitem{lee2012:_proximal}
J.~Lee, Y.~Sun, and M.~Saunders, ``Proximal newton-type methods for convex
  optimization,'' in \emph{Advances in Neural Information Processing Systems},
  Lake Tahoe, NV, Dec. 2012, pp. 836--844.

\bibitem{becker12:_newton}
S.~Becker and J.~Fadili, ``A quasi-newton proximal splitting method,'' in
  \emph{Advances in Neural Information Processing Systems}, Lake Tahoe, NV,
  Dec. 2012, pp. 2618--2626.

\bibitem{thi05:_dc_differ_convex_funct_progr}
H.~A. {Le~Thi} and T.~{Pham~Dinh}, ``The dc (difference of convex functions)
  programming and dca revisited with dc models of real world nonconvex
  optimization problems,'' \emph{Annals of Operations Research}, vol. 133, no.
  1-4, pp. 23--46, 2005.

\bibitem{dinh97:_convex_analy_approac_dc_progr}
T.~{Pham~Dinh} and H.~A. {Le~Thi}, ``Convex analysis approach to dc
  programming: Theory, algorithms and applications,'' \emph{Acta Mathematica
  Vietnamica}, vol.~22, no.~1, pp. 287--355, 1997.

\bibitem{akoa08:_combin_dc_algor_dcas_decom}
F.~Akoa, ``Combining dc algorithms (dcas) and decomposition techniques for the
  training of nonpositive semidefinite kernels,'' \emph{Neural Networks, IEEE
  Transactions on}, vol.~19, no.~11, pp. 1854--1872, Nov 2008.

\bibitem{jenatton10:_proxim_method_for_spars_hierar_diction_learn}
R.~Jenatton, J.~Mairal, G.~Obozinski, and F.~Bach, ``Proximal methods for
  sparse hierarchical dictionary learning,'' in \emph{Proceedings of
  International Conference on Machine Learning}, Tel Aviv, Israel, Jun. 2010,
  pp. 487--494.

\bibitem{rakotomamonjy12:_direc}
A.~Rakotomamonjy, ``Direct optimization of the dictionary learning problem,''
  \emph{IEEE Trans. on Signal Processing}, vol.~61, no.~12, pp. 5495--5506,
  2013.

\bibitem{srebro05:_maxim_margin_matrix_factor}
N.~Srebro, J.~Rennie, and T.~S. Jaakkola, ``Maximum-margin matrix
  factorization,'' in \emph{Advances in neural information processing systems},
  no. Vancouver, BC, Dec., 2004, pp. 1329--1336.

\bibitem{ertekin11:_noncon}
S.~Ertekin, L.~Bottou, and C.~Giles, ``Nonconvex online support vector
  machines,'' \emph{IEEE Trans. on Pattern Analysis and Machine Intelligence},
  vol.~33, no.~2, pp. 368--381, 2011.

\bibitem{collobert:2006}
R.~Collobert, F.~Sinz, J.~Weston, and L.~Bottou, ``Trading convexity for
  scalability,'' in \emph{Proceedings of the Twenty-third International
  Conference on Machine Learning (ICML 2006)}.\hskip 1em plus 0.5em minus
  0.4em\relax ACM Press, 2006, pp. 201--208.

\bibitem{Yuille_CCCP_2001}
A.~L. Yuille, A.~Rangarajan, and A.~Yuille, ``The concave-convex procedure
  (cccp),'' \emph{Advances in neural information processing systems}, vol.~2,
  pp. 1033--1040, Vancouver, BC, Dec. 2002.

\bibitem{courty2014domain}
N.~Courty, R.~Flamary, and D.~Tuia, ``Domain adaptation with regularized
  optimal transport,'' in \emph{Machine Learning and Knowledge Discovery in
  Databases}.\hskip 1em plus 0.5em minus 0.4em\relax Springer, Nancy, France,
  Sep. 2014, pp. 274--289.

\bibitem{jenatton10:_struc_spars_princ_compon_analy}
R.~Jenatton, G.~Obozinski, and F.~Bach, ``Structured sparse principal component
  analysis.'' in \emph{Proceedings of the International Conference on
  Artificial Intelligence and Statistics}, Y.~W. Teh and D.~M. Titterington,
  Eds., vol.~9, Chia, Italy, May 2010, pp. 366--373.

\bibitem{richard2012estimation}
E.~Richard, P.-A. Savalle, and N.~Vayatis, ``Estimation of simultaneously
  sparse and low rank matrices.'' in \emph{Proceedings of the International
  Conference in Machine Learning}.\hskip 1em plus 0.5em minus 0.4em\relax
  Omnipress, Edinburgh, Scotland, Jun. 2012.

\bibitem{deng2013low}
Y.~Deng, Q.~Dai, R.~Liu, Z.~Zhang, and S.~Hu, ``Low-rank structure learning via
  nonconvex heuristic recovery,'' \emph{Neural Networks and Learning Systems,
  IEEE Transactions on}, vol.~24, no.~3, pp. 383--396, 2013.

\bibitem{Zhong_nips}
K.~Zhong, E.-H. Yen, I.~S. Dhillon, and P.~K. Ravikumar, ``Proximal
  quasi-newton for computationally intensive l1-regularized m-estimators,'' in
  \emph{Advances in Neural Information Processing Systems 27}, Z.~Ghahramani,
  M.~Welling, C.~Cortes, N.~Lawrence, and K.~Weinberger, Eds.\hskip 1em plus
  0.5em minus 0.4em\relax Curran Associates, Inc., Montreal, Canada, Dec. 2014,
  pp. 2375--2383.

\bibitem{FigueiredoNowakGradProj2007}
M.~Figueiredo, R.~Nowak, and S.~Wright, ``Gradient projection for sparse
  reconstruction: application to compressed sensing and other inverse
  problems,'' \emph{IEEE Journal of Selected Topics in Signal Processing:
  Special Issue on Convex Optimization Methods for Signal Processing}, vol.~1,
  no.~4, pp. 586--598, 2007.

\bibitem{golub1996matrix}
G.~Golub and C.~Van~Loan, \emph{Matrix computations}.\hskip 1em plus 0.5em
  minus 0.4em\relax Johns Hopkins University Press, 1996, vol.~3.

\bibitem{gong2013jieping}
P.~Gong, C.~Zhang, Z.~Lu, J.~Huang, and Y.~Jieping, ``A general iterative
  shrinkage and thresholding algorithm for non-convex regularized optimization
  problems,'' in \emph{Proceedings of the 30th International Conference on
  Machine Learning}, Atlanta, Georgia, Jun. 2013, pp. 37--45.

\bibitem{lu12:_sequen_convex_progr_method_class}
Z.~Lu, ``Sequential convex programming methods for a class of structured
  nonlinear programming,'' ArXiv:1210.3039, Tech. Rep., 2012.

\bibitem{PoLing_NIPS2013}
P.-L. Loh and M.~J. Wainwright, ``Regularized m-estimators with nonconvexity:
  Statistical and algorithmic theory for local optima,'' in \emph{Advances in
  Neural Information Processing Systems 26}, C.~Burges, L.~Bottou, M.~Welling,
  Z.~Ghahramani, and K.~Weinberger, Eds., Lake Tahoe, NV, Dec 2013, pp.
  476--484.

\bibitem{dinh98:_dc_optim_algor_solvin_trust_region_subpr}
T.~{Pham~Dinh} and H.~A. {Le~Thi}, ``Dc optimization algorithms for solving the
  trust region subproblem,'' \emph{SIAM Journal of Optimization}, vol.~8, pp.
  476--505, 1998.

\bibitem{tsvm_collobert_jmlr_2006}
R.~Collobert, F.~Sinz, J.~Weston, and L.~Bottou, ``Large scale transductive
  svms,'' \emph{Journal of Machine Learning Research}, vol.~7, pp. 1687--1712,
  2006.

\bibitem{mine81}
H.~Mine and M.~Fukushima, ``A minimization method for the sum of a convex
  function and a continuously differentiable function,'' \emph{Journal of
  Optimization Theory and Applications}, vol.~33, no.~1, pp. 9--23, 1981.

\bibitem{chouzenoux13:_variab_metric_forwar_backw}
E.~Chouzenoux, J.-C. Pesquet, and A.~Repetti, ``Variable metric
  forward--backward algorithm for minimizing the sum of a differentiable
  function and a convex function,'' \emph{Journal of Optimization Theory and
  Applications}, vol. 162, no.~1, pp. 107--132, 2014.

\bibitem{sra11}
S.~Sra, ``Scalable nonconvex inexact proximal splitting,'' in \emph{Advances in
  Neural Information Processing Systems}, Lake Tahoe, NV, Dec. 2012, pp.
  530--538.

\bibitem{attouch2010proximal}
H.~Attouch, J.~Bolte, P.~Redont, and A.~Soubeyran, ``Proximal alternating
  minimization and projection methods for nonconvex problems: an approach based
  on the {K}urdyka-{L}ojasiewicz inequality,'' \emph{Mathematics of Operations
  Research}, vol.~35, no.~2, pp. 438--457, 2010.

\bibitem{attouch2013convergence}
H.~Attouch, J.~Bolte, and B.~F. Svaiter, ``Convergence of descent methods for
  semi-algebraic and tame problems: proximal algorithms, forward--backward
  splitting, and regularized gauss--seidel methods,'' \emph{Mathematical
  Programming}, vol. 137, no. 1-2, pp. 91--129, 2013.

\bibitem{bolte2013proximal}
J.~Bolte, S.~Sabach, and M.~Teboulle, ``Proximal alternating linearized
  minimization for nonconvex and nonsmooth problems,'' \emph{Mathematical
  Programming}, vol. 146, no. 1-2, pp. 459--494, 2014.

\bibitem{zhang10:_analy_multi_convex_relax_spars_regul}
T.~Zhang, ``Analysis of multi-stage convex relaxation for sparse
  regularization,'' \emph{Journal of Machine Learning Researc}, vol.~11, pp.
  1081--1107, 2010.

\bibitem{boisbunon2014active}
A.~Boisbunon, R.~Flamary, and A.~Rakotomamonjy, ``Active set strategy for
  high-dimensional non-convex sparse optimization problems,'' in
  \emph{Acoustics, Speech and Signal Processing (ICASSP), IEEE International
  Conference on}.\hskip 1em plus 0.5em minus 0.4em\relax IEEE, Firenze, Italy,
  May 2014, pp. 1517--1521.

\bibitem{rakotomamonjy11:_ell_p_ell_q_penal}
A.~Rakotomamonjy, R.~Flamary, G.~Gasso, and S.~Canu, ``$\ell_p-\ell_q$ penalty
  for sparse linear and sparse multiple kernel multi-task learning,,''
  \emph{IEEE Trans. on Neural Networks}, vol.~22, no.~8, pp. 1307--1320, 2011.

\bibitem{chapelle05:_semi}
O.~Chapelle and A.~Zien, ``Semi-supervised classification by low density
  separation,'' in \emph{Proceedings of the Tenth International Workshop on
  Artificial Intelligence and Statistic}, Barbados, Jan. 2005, pp. 57--64.

\bibitem{joachims_transd}
T.~Joachims, ``Transductive inference for text classification using svms,'' in
  \emph{Proceedings of The 16th International Conference on Machine Learning},
  vol.~99, Bled, Slovenia, Jun. 1999, pp. 200--209.

\end{thebibliography}

\end{document}